%% file: ste-ns-icml-camera-0617.tex
\documentclass[nohyperref]{article}
\usepackage{microtype}
\usepackage{graphicx}
\usepackage{subfigure}
\usepackage{booktabs} 

\usepackage{hyperref}

\usepackage[accepted]{icml2022}
 
\usepackage{amsmath}
\usepackage{amssymb}
\usepackage{mathtools}
\usepackage{amsthm}

\usepackage[capitalize,noabbrev]{cleveref}

\usepackage[normalem]{ulem}

\theoremstyle{plain}
\newtheorem{theorem}{Theorem}[section]
\newtheorem{proposition}[theorem]{Proposition}

\newtheorem{example}[theorem]{Example}
\theoremstyle{definition}

\theoremstyle{remark}

\icmltitlerunning{Injecting Logical Constraints into Neural Networks via Straight-Through Estimators}

\usepackage{multirow}
\usepackage{bbm}

\def\bi{\begin{itemize}}
	
	\def\ei{\end{itemize}}
\def\beq{\begin{equation}}
	\def\eeq#1{\label{#1}\end{equation}}
\def\ba{\begin{array}}
	\def\ea{\end{array}}

\def\wt{\widetilde}		
\def\xx{\mathbf{x}}

\def\false{\hbox{\sc false}}
\def\true{\hbox{\sc true}}

\long\def\BOC#1\EOC{\message{(Commented text )}}
\long\def\BOCC#1\EOCC{\message{(Commented text )}}
\long\def\BOCCC#1\EOCCC{\message{(Commented text )}}

\long\def\NBB#1{}

\begin{document}
	
\twocolumn[
\icmltitle{Injecting Logical Constraints into Neural Networks \\ via Straight-Through Estimators} 
	
\icmlsetsymbol{equal}{*}
	
\begin{icmlauthorlist}
\icmlauthor{Zhun Yang}{asu}
\icmlauthor{Joohyung Lee}{asu,sr}
\icmlauthor{Chiyoun Park}{sr}
\end{icmlauthorlist}

\icmlaffiliation{sr}{
    Samsung Research,
	Samsung Electronics Co.,  
	Seoul, South Korea}
\icmlaffiliation{asu}{
	School of Computing and Augmented Intelligence,
	Fulton Schools of Engineering, Arizona State University, Tempe, AZ, USA}
	
\icmlcorrespondingauthor{Joohyung Lee}{joolee@asu.edu}
	
\icmlkeywords{neuro-symbolic learning, logical reasoning, constraint satisfaction}
	
\vskip 0.3in
]
	
\printAffiliationsAndNotice{}
	
\begin{abstract}
Injecting discrete logical constraints into neural network learning is one of the main challenges in neuro-symbolic AI. We find that a straight-through-estimator, a method introduced to train binary neural networks, could effectively be applied to incorporate logical constraints into neural network learning. More specifically, we design a systematic way to represent discrete logical constraints as a loss function; minimizing this loss using gradient descent via a straight-through-estimator updates the neural network's weights in the direction that the binarized outputs satisfy the logical constraints. The experimental results show that by leveraging GPUs and batch training, this method scales significantly better than existing neuro-symbolic methods that require heavy symbolic computation for computing gradients. Also, we demonstrate that our method applies to different types of neural networks, such as MLP, CNN, and GNN, making them learn  with no or fewer labeled data by learning directly from known constraints.
\end{abstract}

\section{Introduction}\label{sec:intro}
	
Neuro-symbolic AI \cite{besold17neural,mao19neuro,deraedt19neuro,garcez19neural} aims to combine deep neural network learning and symbolic AI reasoning, which look intrinsically different from each other on the surface. 
It appears hard to incorporate discrete logical reasoning into the conventional gradient descent method that deals with continuous values. 
Some recent works in neuro-symbolic AI \cite{manhaeve18deepproblog,yang20neurasp,pogancic20differentiation,tsamoura21neural} associate continuous parameters in neural networks (NNs) with logic languages so that logical reasoning applied to NN outputs produces ``semantic loss'' \cite{xu18asemantic}. Minimizing such loss leads to updating NN parameters via backpropagation through logic layers. Like human learning that leverages known constraints, these methods have shown promising results that allow NNs to learn effectively with fewer data leveraging the semantic constraints.
On the other hand, the symbolic computation performed during backpropagation is implemented by weighted model counting using circuits \cite{darwiche11sdd,manhaeve18deepproblog,tsamoura21neural} or by calling symbolic solvers \cite{poganvcic19differentiation,yang20neurasp}, which are often computationally expensive; it takes too long to generate arithmetic circuits
or enumerate all models or proofs by calling symbolic solvers. 
	
One main reason for the development of the different ideas is that a naive representation of discrete constraints as a loss function is not meaningfully differentiable. Even for the intervals that it is differentiable, the gradient is zero, so NNs won't update their weights. To address this, we turn to the idea of straight-through estimators (STE) \cite{courbariaux15binaryconnect}, which were originally introduced to train binary neural networks (BNNs) --- neural networks with binary weights and activation at run-time. 
The main idea of STE is to use a binarization function in forward propagation while its gradient, which is zero almost everywhere, is replaced by the gradient of a different, meaningfully differentiable function in backward propagation.
It turns out that the method works well for NN quantization in practice. 
	
However, adopting STE alone is not enough for learning with constraints.
We need a systematic method of encoding logical constraints as a loss function and ensure that its gradient enables NNs to learn logical constraints.
	
This paper makes the following contributions.
\begin{itemize}
\item  We design a systematic way to encode logical constraints in propositional logic as a loss function in neural network learning, which we call {\em CL-STE}. We demonstrate that minimizing this loss function via STE enforces the logical constraints in neural network learning so that neural networks learn from the explicit constraints. 
				
\item We show that by leveraging GPUs and batch training, CL-STE scales significantly better than the other neuro-symbolic learning methods that use heavy symbolic computation for computing gradients. 
				
\item We also find that the concept of Training Gate Function (TGF) \cite{kim20plug}, which was applied to channel pruning, 
is closely related to STE. We establish the precise relationship between them, which gives a new perspective of STE. 
\end{itemize}
		
The paper is organized as follows. Section~\ref{sec:related} presents related works, and Section~\ref{sec:ste-tgf} reviews STE and TFG and establish their relationships.
Section~\ref{sec:ste-learning} presents our loss function representation of logical constraints and proves its properties assuming minimization via STE, and Section~\ref{sec:experiments} shows experimental results. 

The implementation of our method is publicly available online at 
\url{https://github.com/azreasoners/cl-ste}.
 
\section{Related Work} \label{sec:related}

Our work is closely related to \cite{xu18asemantic}, which proposes a semantic loss function to bridge NN outputs and logical constraints. The method treats an NN output as probability and computes semantic loss as the negative logarithm of the probability to generate a state satisfying the logical constraints. 
Their experiments show that the encoded semantic loss function guides the learner to achieve state-of-the-art results in supervised and semi-supervised learning on multi-class classification. For the efficient computation of a loss function, they encode logical constraints in Sentential Decision Diagram (SDD) \cite{darwiche11sdd}. However, generating SDDs is computationally expensive for most practical tasks. 
	
Several neuro-symbolic formalisms, such as DeepProbLog \cite{manhaeve18deepproblog}, NeurASP \cite{yang20neurasp}, and  NeuroLog \cite{tsamoura21neural}, have been proposed to integrate neural networks with logic programming languages. 
Since discrete logical inference cannot be in general captured via a differentiable function, they use relaxation to weighted models or probability. 
While this approach provides a systematic representation of constraints, the symbolic computation is often the bottleneck in training.

Since fuzzy logic operations are naturally differentiable, several works, such as Logic Tensor Network \cite{serafini2016logic}, Continuous Query Decomposition \cite{arakelyan20complex}, Semantic Based Regularization \cite{diligenti17semantic,roychowdhury21regularizing}, directly apply fuzzy operators to neural network outputs.
However, as stated in~\cite{marra21statistical}, the fuzzification procedure alters the logical properties of the original theory (such as satisfiability).
	
Other works train neural networks for learning satisfiability, such as \cite{wang19satnet,selsam19learning}. 
SATNet \cite{wang19satnet} builds on a line of research exploring SDP relaxations as a tool for solving MAXSAT, producing tighter approximation guarantees than standard linear programming relaxation. 

Graph Neural Networks (GNNs) \cite{battaglia18relational,lamb20graph} have been widely applied for logical reasoning. For example, Recurrent Relational Network (RRN) was able to learn how to solve Sudoku puzzles. 
GNNs use message-passing to propagate logical constraints in neural networks, but they do not have the mechanism to specify the logical constraints directly as we do.

While STE has not been exploited in neuro-symbolic learning to our best knowledge, 
\cite{pogancic20differentiation}'s work 
is related in that it also uses a gradient that is different from the forward function's gradient.  It uses the gradient obtained from a linear relaxation of the forward function.
The work also requires a combinatorial solver to compute the gradient. 

\section{Straight-Through-Estimators and Trainable Gate Function} \label{sec:ste-tgf}

{\bf Review}.\ \ \ 
STEs are used to estimate the gradients of a discrete function. \citeauthor{courbariaux15binaryconnect} (\citeyear{courbariaux15binaryconnect}) consider a binarization function~$b$ that transforms real-valued weights $x$ into discrete values $b(x)$ as $b(x)=1$ if $x\ge 0$ and $b(x)=0$ otherwise.
A loss function $L$ is defined on binarized weights $b(x)$, but the gradient descent won't update binarized weights in small increments. However, using STE, we could update the real-valued weights $x$ that are input to $b(x)$. In the end, a quantized model consists of binarized weights $b(x)$ only. 
More specifically, according to the chain rule, the gradient of loss $L$ w.r.t.~$x$ is
$
\frac{\partial L}{\partial x} = \frac{\partial L}{\partial b(x)}\times \frac{\partial b(x)}{\partial x}
$, 
where $\frac{\partial b(x)}{\partial x}$ is zero almost everywhere.
The idea is to replace $\frac{\partial b(x)}{\partial x}$ with 
an STE
$\frac{\partial s(x)}{\partial x}$ for some (sub)differentiable function $s(x)$. The STE $\frac{\partial s(x)}{\partial x}$ is called the {\em identity STE} (iSTE) if $s(x)=x$ and is called the {\em saturated STE} (sSTE) if $s(x)=clip(x, [-1,1]) = min(max(x, -1), 1)$.
Since $\frac{\partial s(x)}{\partial x}=1$, by $\frac{\partial L}{\partial x} \stackrel{iSTE}{\approx} \frac{\partial L}{\partial b(x)}$, we denote the identification of $\frac{\partial L}{\partial x}$ with $\frac{\partial L}{\partial b(x)}$ under iSTE.

The binarization function $b(x)$ passes only the sign of $x$ while information about the magnitude of $x$ is lost \cite{simons19review}.
In XNOR-Net~\cite{rastegari16xnor}, the input $x$ is normalized to have the zero mean and a small variance before the binarization to reduce the information loss. In this paper, 
we normalize $x$ by turning it into a probability using softmax or sigmoid activation functions. Indeed, several neuro-symbolic learning methods (e.g., DeepProbLog, NeurASP, NeuroLog) assume the neural network outputs fed into the logic layer are normalized as probabilities. 
To address a probabilistic input, we introduce a variant binarization function $b_p(x)$ for probabilities $x\in \left[0,1\right]$:
$b_p(x) = 1$ if $x\ge 0.5$ and $b_p(x)=0$ otherwise.
It is easy to see that iSTE and sSTE work the same with $b_p(x)$ since $x=clip(x,[-1,1])$ when $x\in [0,1]$. 
A vector $\mathbf{x}$ is allowed as input to the binarization functions $b$ and $b_p$, in which case they are applied to each element of $\mathbf{x}$. 

\begin{figure}
	\centerline{\includegraphics[width=1\columnwidth]{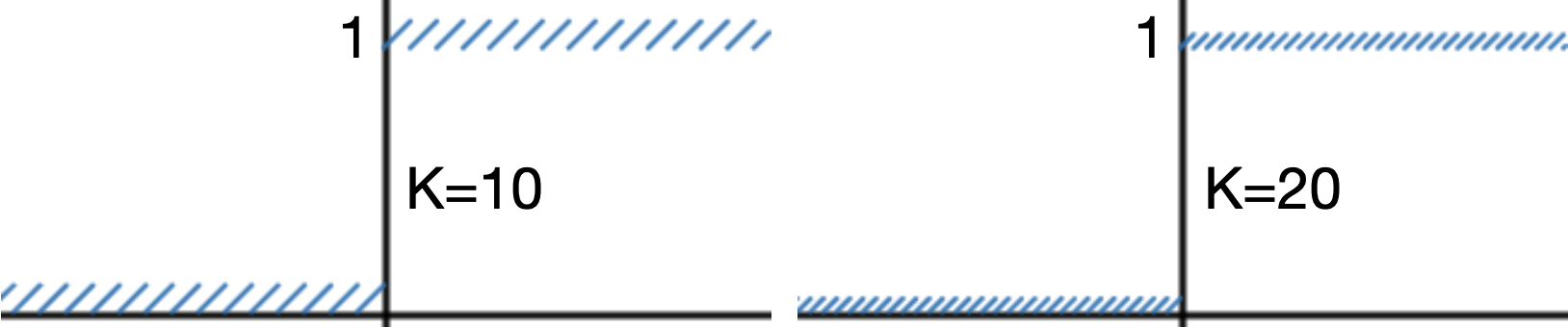}}
	\caption{Trainable gate function $\wt{b}^K(x)$ when $g(x)=1$}
	\label{fig:diode_bx}
\end{figure}

{\bf TGF and Its Relation to STE}.\ \ \ 
The concept of STE is closely related to that of the Trainable Gate Function (TGF) from~\cite{kim20plug}, which was applied to channel pruning.
Instead of replacing the gradient $\frac{\partial b(x)}{\partial x}$ with an STE, TGF tweaks the binarization function $b(x)$ to make it meaningfully differentiable. More specifically, a differentiable binarization function $\widetilde{b}^K$ is defined as
\begin{align} \label{eq:b^K}
	\wt{b}^K(x) = b(x) + s^K(x)g(x),
\end{align}
where $K$ is a large constant; $s^K(x) = \frac{Kx - \lfloor{Kx}\rfloor}{K}$ is called a {\em gradient tweaking} function, whose value is less than $\frac{1}{K}$ and whose gradient is always 1 wherever differentiable;  $g(x)$ is called a {\em gradient shaping} function, which could be an arbitrary function, but the authors note that the selection does not affect the results critically and $g(x)=1$ can be adopted without significant loss of accuracy.
As obvious from Figure~\ref{fig:diode_bx}, as $K$ becomes large, TGF $\wt{b}^K(x)$ is an approximation of $b(x)$, but its gradient is $1$ wherever differentiable. 

Proposition~\ref{prop:diode} tells us a precise relationship between TGF and STE:  when $K$ is big enough, the binarization function $b(x)$ with iSTE or sSTE can be simulated by TGF.
In other words, Proposition~\ref{prop:diode} allows us to visualize $b(x)$ with STE as the TGF $\wt{b}^K(x)$ with $K=\infty$ as  Figure~\ref{fig:diode_bx} illustrates.

\begin{proposition} \label{prop:diode}
	When $K$ approaches $\infty$ and $|g(x)|\leq c$ for some constant $c$, the value of $\wt{b}^K(x)$ converges to $b(x)$:
	\begin{align*}
		\lim_{K \to \infty}\wt{b}^K(x) = b(x). 
	\end{align*}
	The gradient $\frac{\partial \wt{b}^K(x)}{\partial x}$, wherever defined, is exactly the iSTE of $\frac{\partial b(x)}{\partial x}$ if $g(x)=1$, or the sSTE of $\frac{\partial b(x)}{\partial x}$ if 
	\begin{align*}
		g(x) = \begin{cases}
			1 & \text{if $-1 \leq x \leq 1$}, \\
			0 & \text{otherwise.}
		\end{cases}
	\end{align*}
\end{proposition}
Proposition~\ref{prop:diode} still holds if we replace $b(x)$ with $b_p(x)$.

The proposition yields insights into STE and TGF in terms of each other. As shown in Figure~\ref{fig:diode_bx}, TGF is a sawtooth function that approximates a step function as $K$ becomes large. At large, TGF works like a discrete function, but it is differentiable almost everywhere. In view of Proposition~\ref{prop:diode}, this fact gives an idea why the STE method works in practice.
On the other hand, the proposition tells that the implementation of TGF can be replaced with STE. That could be better because TGF in equation~\eqref{eq:b^K} requires that $K$ approximate infinity and be non-differentiable when $x$ is a multiple of $\frac{1}{K}$ whereas STE is differentiable at every $x$. 

\section{Enforcing Logical Constraints using STE} \label{sec:ste-learning}

This section presents our method of encoding logical constraints in propositional logic as a loss function so that minimizing its value via STE makes neural network prediction follow the logical constraints. 

\subsection{Encoding CNF as a Loss Function Using STE}

We first review the terminology in propositional logic. A {\em signature} is a set of symbols called {\em atoms}. 
Each atom represents a proposition that is true or false.
A {\em literal} is either an atom $p$ ({\em positive literal}) or its negation $\neg p$ ({\em negative literal}). 
A {\em clause} is a disjunction over literals, e.g., $p_1 \lor \neg p_2 \lor p_3$.
A {\em Horn clause} is a clause with at most one positive literal.
We assume a {\em (propositional) theory} consisting of a set of clauses (sometimes called a {\em CNF (Conjunctive Normal Form) theory}). 
A truth assignment to atoms {\em satisfies} (denoted by $\models$) a theory if at least one literal in each clause is true under the assignment.
A theory is {\em satisfiable} if at least one truth assignment satisfies the theory. 
A theory {\em entails} (also denoted by $\models$) a literal if every truth assignment that satisfies the theory also satisfies that literal. 

We define a general loss function $L_{cnf}$ for any CNF theory as follows.
Here, bold upper and lower letters (e.g., $\mathbf{C}$ and $\mathbf{v}$) denote matrices and vectors, respectively; $\mathbf{C}[i,j]$ and $\mathbf{v}[i]$ denote their elements. 

Consider a propositional signature $\sigma = \{p_1, \dots, p_n\}$. Given 
(i) a theory $C$ consisting of $m$ clauses (encoding domain knowledge),  
(ii) a set $F$ of atoms denoting some atomic facts that we assume known to be true (representing the ground-truth label of a data instance), and 
(iii) a truth assignment $v$ such that $v\models F$, we construct their matrix/vector representations as
\begin{itemize}
\item the matrix $\mathbf{C} \in \{-1,0,1 \}^{m\times n}$ to represent the theory such that $\mathbf{C}[i,j]$ is $1$ ($-1$, resp.) if $p_j$ ($\neg p_j$, resp.) belongs to the $i$-th clause in the theory, and is $0$ if neither $p_j$ nor $\neg p_j$ belongs to the clause;
\item the vector $\mathbf{f}\in \{0,1\}^{n}$ to represent $F$ such that $\mathbf{f}[j]$ is $1$ if $p_j\in F$ and is $0$ otherwise; and
\item the vector $\mathbf{v}\in \{0,1\}^{n}$ to represent $v$ such that $\mathbf{v}[j]$ is $1$ if $v(p_j)=\true$, and is $0$ if $v(p_j)=\false$.
\end{itemize}

\begin{figure}
\centerline{\includegraphics[width=0.9\columnwidth]{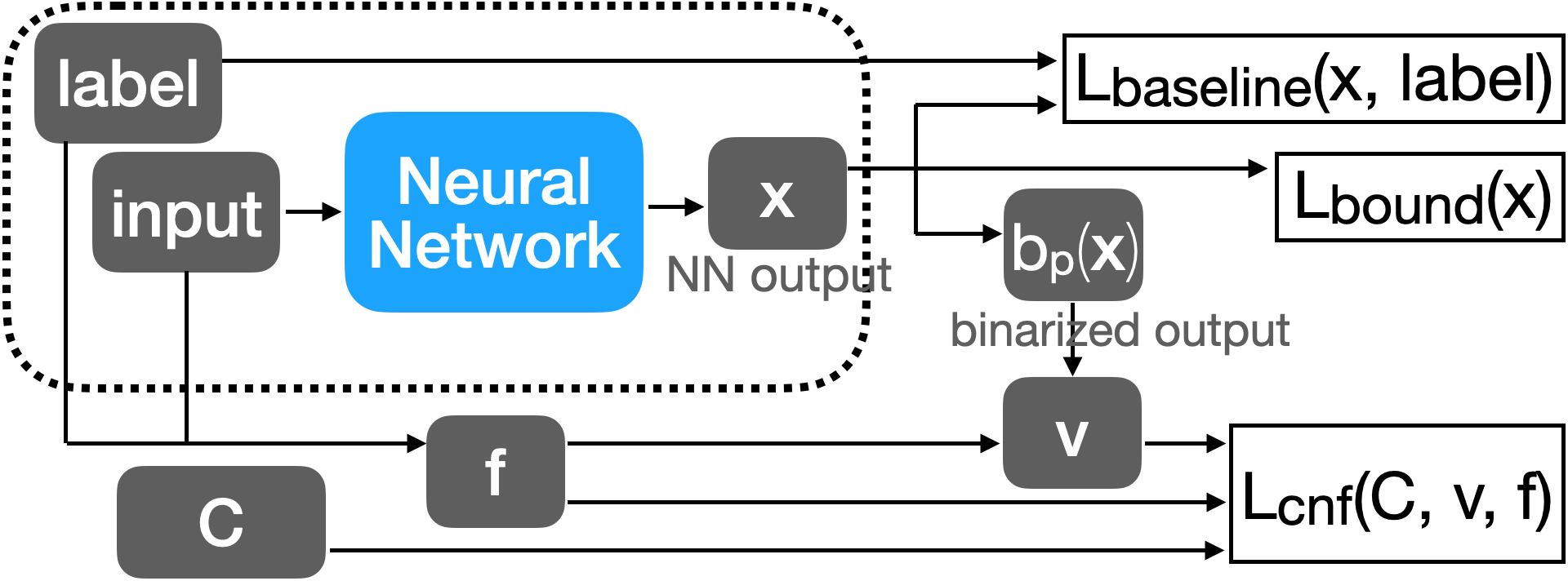}}
\caption{Architecture that overlays constraint loss} 
\label{fig:archi-binary}
\end{figure}

Figure~\ref{fig:archi-binary} shows an architecture that overlays the two loss functions $L_{bound}$ and $L_{cnf}$ over the neural network output, where $L_{cnf}$ is the main loss function to encode logical constraints and $L_{bound}$ is a regularizer to limit the raw neural network output not to grow too big (more details will follow).
The part $\mathbf{input}$ is a tensor (e.g., images) for a data instance;
$\mathbf{label}$ denotes the labels of input data;
$\mathbf{C}$ encodes the domain knowledge, 
$\xx \in [0,1]^n$ denotes the NN output (in probability), and $\mathbf{f}\in \{0,1\}^{n}$ records the known facts in that data instance ({e.g., given digits in Sudoku}).\footnote{
In case the length of $\xx$ is less than $n$, we pad $\xx$ with $0$s for all the atoms that are not related to NN output.
}
Let $\mathbbm{1}_{\{k \}}(X)$ denote an indicator function that replaces every element in $X$ with $1$ if it is $k$ and with $0$ otherwise.
Then the {binary} prediction $\mathbf{v}$ is constructed as
$\mathbf{v} = \mathbf{f} + \mathbbm{1}_{\{0 \}}(\mathbf{f}) \odot b_p(\xx)$, where $\odot$ denotes element-wise multiplication.
Intuitively, $\mathbf{v}$ is the binarized NN output with part of it strictly following the given facts specified in $\mathbf{f}$ (ensuring $v\models F$).

\begin{example} \label{ex:mnistAdd}
Consider a simple example {\bf mnistAdd} from \cite{manhaeve18deepproblog}, where the task is, given a pair of MNIST digit images and their sum as the label, to let a neural network learn the digit classification of the input images. The example is used to demonstrate how NNs can learn from known constraints.  
In Figure~\ref{fig:archi-binary}, the input consists of two-digit images $i_1$ and $i_2$, and the label is an integer $l$ in $\{0, ..., 18\}$ denoting the sum of $i_1$ and $i_2$.
The neural network is the same Convolutional Neural Network (CNN)
used in \cite{manhaeve18deepproblog}. 

The theory for this problem consists of the following clause for $l \in \{0,\dots,18\}$,
where $sum(l)$ represents ``the sum of $i_1$ and $i_2$ is $l$'' and $pred(n_1, n_2)$ represents ``the neural network predicts $i_1$ and $i_2$ as $n_1$ and $n_2$ respectively'':
\begin{align*}
\neg sum(l) \lor \bigvee\limits_{\substack{n_1,n_2\in \{0,\dots,9 \}:\\n_1+n_2=l}} pred(n_1, n_2).
\end{align*}
This theory contains
$19 + 100 = 119$ atoms for $sum/1$ and $pred/2$ respectively. We construct the matrix $\mathbf{C} \in \{-1,0,1 \}^{19\times 119}$, where each row represents a clause. For instance, the row for the clause $\neg sum(1) \lor pred(0, 1) \lor pred(1, 0)$ is a vector in $\{-1,0,1\}^{1\times 119}$ containing a single $-1$ for atom $sum(1)$, two $1$s for atoms $pred(0, 1)$ and $pred(1, 0)$, and 116 $0$s. 

Vectors $\mathbf{f}$ and $\mathbf{v}$ are in $\{0,1\}^{119}$ constructed from each data instance $\langle i_1,i_2,l \rangle$.
The fact vector $\mathbf{f}$ contains a single $1$ for atom $sum(l)$ (ground truth label) and 118 $0$s.  
To obtain the prediction vector $\mathbf{v}$, we 
(i) feed images $i_1$,$i_2$ into the CNN (with softmax at the last layer) from \cite{manhaeve18deepproblog} to obtain the outputs 
{
$\xx_1, \xx_2 \in [0,1]^{10}$ 
}
(consisting of probabilities),
(ii) construct the vector 
{
$\mathbf{x} \in [0,1]^{100}$
}
(for 100 atoms of $pred/2$) such that
$\xx[10i+j]$ is $\xx_1[i] \times \xx_2[j]$ for $i,j\in \{0,\dots,9\}$,
(iii) update $\mathbf{x}$ as the concatenation of $\xx$ and $\{0\}^{19}$, and
(iv) finally, let $\mathbf{v} = \mathbf{f} + \mathbbm{1}_{\{0 \}}(\mathbf{f}) \odot b_p(\xx)$.
\end{example}

\smallskip
Using $\mathbf{C}$, $\mathbf{v}$, and $\mathbf{f}$, 
we define the {\em CNF loss} $L_{cnf}(\mathbf{C}, \mathbf{v}, \mathbf{f})$ as follows:
\begin{align}
\mathbf{L}_f =\ & \mathbf{C} \odot \mathbf{f}  \label{eq:lg} \\
\mathbf{L}_v =\ & \mathbbm{1}_{\{1\}}(\mathbf{C}) \odot \mathbf{v} + \mathbbm{1}_{\{-1\}}(\mathbf{C}) \odot (1-\mathbf{v})  \label{eq:lv} \\
\mathbf{deduce} =\ & \mathbbm{1}_{\{1\}}\Big(sum(
{ 
\mathbf{C} \odot \mathbf{C}
}
) - sum(\mathbbm{1}_{\{-1\}}(\mathbf{L}_f)) \Big)  \label{eq:up} \\
\mathbf{unsat} =\ & prod(1-\mathbf{L}_v) \label{eq:unsat} \\
\mathbf{keep} =\ & sum(\mathbbm{1}_{\{1\}}(\mathbf{L}_v) \odot (1-\mathbf{L}_v) + \mathbbm{1}_{\{0\}}(\mathbf{L}_v) \odot \mathbf{L}_v) \label{eq:keep} \\
L_{deduce} =\ & sum(\mathbf{deduce} \odot \mathbf{unsat}) \label{eq:lup} \\
L_{unsat} =\ & avg( \mathbbm{1}_{\{1\}}(\mathbf{unsat}) \odot \mathbf{unsat} )  \label{eq:lunsat} \\
L_{sat} =\ & avg(\mathbbm{1}_{\{0\}}(\mathbf{unsat}) \odot \mathbf{keep} ) \label{eq:lsat}  
\end{align}
\vspace{-2em}
\begin{align}
L_{cnf}(\mathbf{C}, \mathbf{v}, \mathbf{f}) =\ & L_{deduce} + L_{unsat} + L_{sat} \label{eq:lcnf}
\end{align}
where $prod(X)$, $sum(X)$, and $avg(X)$ compute the product, sum, and average of the elements in $X$ along its last dimension. \footnote{The aggregated dimension is ``squeezed," which is the default behavior in PyTorch aggregate functions ({\tt keepdim} is False).}
Although these equations may look complex, it helps to know that they use the form  $\mathbbm{1}_{\{k\}}(X_1) \odot X_2$, where the indicator function $\mathbbm{1}_{\{k\}}(X_1)$ can be seen as a constant that is multiplied to a trainable variable $X_2$. Take equation~\eqref{eq:lunsat} as an example. 
To minimize $L_{unsat}$, the NN parameters will be updated towards making $\mathbf{unsat}[i]$ to be $0$ whenever $\mathbbm{1}_{\{1\}}(\mathbf{unsat})$ is $1$, i.e., towards making unsatisfied clauses satisfied. 

In equations~\eqref{eq:lg} and \eqref{eq:lv}, $\mathbf{f}$ and $\mathbf{v}$ are treated as matrices in $\{0,1\}^{1\times n}$ to have element-wise multiplication (with broadcasting) with a matrix in $\{-1,0,1\}^{m\times n}$. Take equation~\eqref{eq:lg} as an example, $\mathbf{L}_f[i,j] = \mathbf{C}[i,j] \times \mathbf{f}[j]$. 
$\mathbf{L}_f$ is the matrix in $\{-1,0,1\}^{m\times n}$ such that (i) $\mathbf{L}_f[i,j]=1$ iff clause $i$ contains literal $p_j$ and $p_j\in F$;  (ii) $\mathbf{L}_f[i,j]=-1$ iff clause $i$ contains literal $\neg p_j$ and $p_j\in F$; (iii) otherwise, $\mathbf{L}_f[i,j]=0$.

$\mathbf{L}_v$ is the matrix in $\{0,1\}^{m\times n}$ such that $\mathbf{L}_v[i,j]=1$ iff clause $i$ contains a literal ($p_j$ or $\neg p_j$) for atom $p_j$ and this literal is $\true$ under $v$.

In equations~\eqref{eq:up}, \eqref{eq:unsat}, and \eqref{eq:keep},
$sum(\mathbf{C} \odot \mathbf{C})$ is a vector in $\mathbb{N}^m$ 
representing in each clause the number of literals, 
and $sum(\mathbbm{1}_{\{-1\}}(\mathbf{L}_f))$ is a vector in $\mathbb{N}^m$ representing in each clause the number of literals that are $\false$ under $F$ (i.e., the number of literals of the form $\neg p$ such that $p\in F$).
Consequently, $\mathbf{deduce}$ is a vector in $\{0,1 \}^m$ where $\mathbf{deduce}[i]$ is $1$ iff clause $i$ has all but one literal being $\false$ under $F$.
If $C\cup F$ is satisfiable and a clause has all but one literal being $\false$ under $F$, then we can safely deduce that the remaining literal is $\true$. For instance, in a clause for Sudoku 
\beq
\neg a(1, 1, 9) \lor \neg a(1,2,9),
\eeq{clause:sudoku}
if $a(1,1,9)$ is in the ground-truth label (i.e., in $F$) but $a(1,2,9)$ is not, we can safely deduce $\neg a(1,2,9)$ is true.
It follows that such a clause is always a Horn clause.
Intuitively, the vector $\mathbf{deduce}$ represents the clauses that such deduction can be applied given $F$.

The vector $\mathbf{unsat}\in \{0,1 \}^{m}$ indicates which clause is not satisfied by the truth assignment $v$, where $\mathbf{unsat}[i]$ is 1 iff $v$ does not satisfy the $i$-th clause.
The vector $\mathbf{keep}\in \{0 \}^{m}$ consists of $m$ zeros while its gradient w.r.t. $\mathbf{v}$ consists of non-zeros. Intuitively, the gradient of $\mathbf{keep}$ tries to keep the current predictions $\mathbf{v}$ in each clause. 

In equations~\eqref{eq:lup}, \eqref{eq:lunsat}, and \eqref{eq:lsat},
$L_{deduce}\in \mathbb{N}$ represents the number of clauses that can deduce a literal given $F$ and are not satisfied by $v$. 
The vector $\mathbbm{1}_{\{1\}}(\mathbf{unsat}) \in \{0,1 \}^m$ (and $\mathbbm{1}_{\{0\}}(\mathbf{unsat})$, resp.) indicates the clauses that are not satisfied (and are satisfied, resp.) by $v$. 
Intuitively, for all clauses, minimizing $L_{unsat}$ makes the neural network change its predictions to decrease the number of unsatisfied clauses. 
In contrast, minimizing $L_{sat}$ makes the neural network more confident in its predictions in the satisfied clauses. 
We use $avg$ instead of $sum$ in equations~\eqref{eq:lunsat} and \eqref{eq:lsat} to ensure that the gradients from $L_{unsat}$ and $L_{sat}$ do not overpower those from $L_{deduce}$.
%
Formal statements of these intuitive explanations follow in the next section.

For any neural network output $\xx$ consisting of probabilities, let $\xx^r$ denote the raw value of $\xx$ before the activation function (e.g., softmax or sigmoid) in the last layer. 
Without restriction, the value $\xx^r$ may vary in a large range when trained with STE. When such an output is fed into softmax or sigmoid, it easily falls into a saturation region of the activation function \cite{tang17train}. To resolve this issue, we include another loss function to bound the scale of $\xx^r$:
\begin{align*}
L_{bound}(\xx) &=  avg( \xx^r \odot \xx^r ) .
\end{align*}

To enforce constraints, we add the weighted sum of 
$L_{cnf}(\mathbf{C}, \mathbf{v}, \mathbf{f})$ and $L_{bound}(\xx)$ to the baseline loss (if any), where the weight of each loss is a hyperparameter.
We call this way of semantic regularization the {\em CL-STE} (Constraint Loss via STE) method.

\noindent{\bf Example~\ref{ex:mnistAdd} Continued.}\ \ \ 
{\sl 
Given the matrix $\mathbf{C}$ for the CNF theory, a data instance $\langle i_1,i_2,l \rangle$, the NN outputs $\xx_1, \xx_2$ for $i_1,i_2$, and the vectors $\mathbf{f}, \mathbf{v}$ as constructed in Example~\ref{ex:mnistAdd}, the total loss function used for {\bf mnistAdd} problem is
\[
{\cal L} = L_{cnf}(\mathbf{C}, \mathbf{v}, \mathbf{f})  + 
\sum\limits_{\xx\in\{\xx_1,\xx_2 \}} {0.1\times} L_{bound}(\xx). 
\]
}

\vspace{-1em}
\subsection{Properties of Constraint Loss and Its Gradients}

Proposition~\ref{prop:value} shows the relation between  $L_{deduce}$, $L_{unsat}$, and $L_{sat}$ components in the constraint loss $L_{cnf}$ and its logical counterpart.

\begin{proposition}\label{prop:value}
Given a theory $C$, a set $F$ of atoms, and a truth assignment $v$ such that $v\models F$, let $\mathbf{C}, \mathbf{f}, \mathbf{v}$ denote their matrix/vector representations, respectively. Let $C_{deduce} \subseteq C$ denote the set of Horn clauses $H$ in $C$ such that
all but one literal in $H$ are of the form $\neg p$ such that $p\in F$. \footnote{This implies that the remaining literal is either an atom or $\neg p$ such that $p\not\in F$.}
Then 
\begin{itemize}
\item the minimum values of $L_{deduce}$, $L_{unsat}$, $L_{sat}$, and $L_{cnf}(\mathbf{C}, \mathbf{v}, \mathbf{f})$ are 0;
\item $v\models C_{deduce}$ iff $L_{deduce}$ is 0;
\item $v\models C$ iff $L_{unsat}$ is 0 iff $L_{cnf}(\mathbf{C}, \mathbf{v}, \mathbf{f})$ is 0.
\end{itemize}
\end{proposition}

Clause~\eqref{clause:sudoku} is an example clause in $C_{deduce}$.
There could be many other ways to design $L_{cnf}(\mathbf{C}, \mathbf{v}, \mathbf{f})$ to satisfy the properties in Proposition~\ref{prop:value}.
{Propositions~\ref{prop:gradient:b_p} and \ref{prop:gradient} below justify} our design choice.

\begin{proposition}\label{prop:gradient:b_p}
Given a theory $C$ with $m$ clauses and $n$ atoms and a set $F$ of atoms such that $C\cup F$ is satisfiable, let $\mathbf{C}, \mathbf{f}$ denote their matrix/vector representations, respectively. 
Given a neural network output {$\mathbf{x}\in [0,1]^{n}$ denoting probabilities},
we construct $\mathbf{v} = \mathbf{f} + \mathbbm{1}_{\{0 \}}(\mathbf{f}) \odot b_p(\xx)$ 
and a truth assignment $v$ such that $v(p_j)=\true$ if $\mathbf{v}[j]$ is $1$, and $v(p_j)=\false$ if $\mathbf{v}[j]$ is $0$. 
Let $C_{deduce} \subseteq C$ denote the set of Horn clauses $H$ in $C$ such that all but one literal in $H$ are of the form $\neg p$  where $p\in F$.
Then, for any $j\in \{1,\dots,n\}$,
\begin{enumerate}
\item if $p_j\in F$, all of $\frac{\partial L_{deduce}}{\partial \xx[j]}$, $\frac{\partial L_{unsat}}{\partial \xx[j]}$, and $\frac{\partial L_{sat}}{\partial \xx[j]}$ are zeros;
\item if $p_j\not \in F$,
\begin{align*}
	\frac{\partial L_{deduce}}{\partial \xx[j]}& \stackrel{iSTE}{\approx}
	\begin{cases}
		-c & \text{if $c>0$ clauses in $C_{deduce}$} \\
		&\text{contain literal $p_j$;}\\
		c & \text{if $c>0$ clauses in $C_{deduce}$}\\
		&\text{contain literal $\neg p_j$;} \\
		0 & \text{otherwise;}
	\end{cases} 
\end{align*}
\begin{align*}
\frac{\partial L_{unsat}}{\partial \xx[j]} &
	\stackrel{iSTE}{\approx}
	\frac{c_2 - c_1}{m} \\
	\frac{\partial L_{sat}}{\partial \xx[j]} &
	\stackrel{iSTE}{\approx}
	\begin{cases}
		-\frac{c_3}{m} & \text{if $v\models p_j$} \\
		\frac{c_3}{m} & \text{if $v\not \models p_j$,} 
	\end{cases}
\end{align*}
where $\stackrel{iSTE}{\approx}$ stands for the equivalence of gradients assuming iSTE;\  $c_1$ (and $c_2$, resp.) is the number of clauses in $C$ that are not satisfied by $v$ and contain $p_j$ (and $\neg p_j$, resp.); $c_3$ is the number of clauses in $C$ that are satisfied by $v$ and contain $p_j$ or $\neg p_j$.
\end{enumerate}
\end{proposition}

{
Intuitively, Proposition~\ref{prop:gradient:b_p} ensures the following properties of the gradient $\frac{\partial L_{cnf}(\mathbf{C}, \mathbf{v}, \mathbf{f})}{\partial \xx[j]}$, which consists of $\frac{\partial L_{deduce}}{\partial \xx[j]}$, $\frac{\partial L_{unsat}}{\partial \xx[j]}$, and $\frac{\partial L_{sat}}{\partial \xx[j]}$.

\smallskip
\noindent{\bf P1.~} 
If we know for sure that $p_j$ is $\true$ ($p_j\in F$), these gradients w.r.t. $\xx[j]$ (real values corresponding to $p_j$) are $0$, so they do not affect the truth value of $p_j$. 

\noindent{\bf P2.~}
Otherwise ($F$ does not tell whether $p_j$ is $\true$),
\begin{enumerate}
\item the gradient $\frac{\partial L_{deduce}}{\partial \xx[j]}$
is negative (positive, resp.) to increase (decrease, resp.) the value of $\xx[j]$ by the gradient descent
if $C\cup F$ entails $p_j$ ($\neg p_j$, resp.);
\item the gradient 
$\frac{\partial L_{unsat}}{\partial \xx[j]}$ is negative (positive resp.) to increase (decrease, resp.) the value of $\xx[j]$ by the gradient descent if, among all unsatisfied clauses, more clauses contain $p_j$ than $\neg p_j$ ($\neg p_j$ than $p_j$, resp.);
\item the gradient
$\frac{\partial L_{sat}}{\partial \xx[j]}$ is negative (positive resp.) to increase (decrease, resp.) the value of $\xx[j]$ by the gradient descent if $v\models p_j$ ($v\not \models p_j$, resp.) and there exist satisfied clauses containing literal $p_j$ or $\neg p_j$.
\end{enumerate}

Intuitively, bullet 1 in {\bf P2} simulates a deduction step, which is always correct, while bullets 2 and 3 simulate two heuristics: ``we tend to believe a literal if more unsatisfied clauses contain this literal than its negation'' and ``we tend to keep our prediction on an atom if many satisfied clauses contain this atom.'' This justifies another property below.

\smallskip
\noindent{\bf P3.~}
The sign of the gradient $\frac{\partial L_{cnf}}{\partial \xx[j]}$ is the same as the sign of $\frac{\partial L_{deduce}}{\partial \xx[j]}$ when the latter gradient is non-zero.
\smallskip 
}

\begin{example}
		Consider the theory $C$ below with $m=2$ clauses and 3 atoms 
		\begin{align*}
			&\neg a \lor \neg b \lor c\\
			&\neg a \lor b
		\end{align*}
		and consider the set of given facts $F=\{a\}$. They are represented by matrix 
		$\mathbf{C}=
		\begin{bmatrix}
			-1 & -1 & 1\\
			-1 & 1 & 0
		\end{bmatrix}
		$
		and vector $\mathbf{f} = [1,0,0]$. 
		Suppose a neural network predicts $\mathbf{x} = [0.3,0.1,0.9]$ as the probabilities of the 3 atoms $\{a,b,c\}$. 
		
		With the above matrix and vectors, we can compute
		\begin{align*}
			b_p(\mathbf{x}) &= [0,0,1], \\
			\mathbf{v} &= \mathbf{f} + \mathbbm{1}_{\{0 \}}(\mathbf{f}) \odot b_p(\xx) = [1,0,1].
		\end{align*}
		From $\mathbf{v}$, we construct the truth assignment $v=\{a=\true, b=\false, c=\true\}$. Clearly, $v$ satisfies the first clause but not the second one.
		Given $F=\{a\}$, we see $C_{deduce}$ is $~\neg a \lor b~$.

		According to Proposition~\ref{prop:gradient:b_p}, 
		\begin{align*}
			\frac{\partial L_{deduce}}{\partial \xx} &\stackrel{iSTE}{\approx} [0,-1,0], \ \ \ \ 
			\frac{\partial L_{unsat}}{\partial \xx} &\stackrel{iSTE}{\approx} [0,-\frac{1}{2},0], \\
			\frac{\partial L_{sat}}{\partial \xx} &\stackrel{iSTE}{\approx} [0,\frac{1}{2},-\frac{1}{2}],
		\end{align*}
		\[
		\frac{\partial L_{cnf}}{\partial \xx} =
		\frac{\partial L_{deduce}}{\partial \xx} + \frac{\partial L_{unsat}}{\partial \xx} + \frac{\partial L_{sat}}{\partial \xx}\stackrel{iSTE}{\approx} [0,-1,-\frac{1}{2}].
		\]

		{
		Intuitively, given $C$, $F$, and the current truth assignment~$v$, 
		({\bf P1}) we know $a$ is $\true$ ($a\in F$) thus no need to update it,
		({\bf P2.1} and {\bf P3}) we know for sure that the prediction for $b$ should be changed to $\true$ by deduction on clause $~\neg a \lor b~$ and the given fact $F=\{a\}$, 
		({\bf P2.3}) we tend to strengthen our belief in $c$ being $\true$ due to the satisfied clause $~\neg a \lor \neg b \lor c~$.
		}
\end{example}

{The proposition also holds with another binarization function $b(x)$.} 
\begin{proposition}\label{prop:gradient}
Proposition~\ref{prop:gradient:b_p} still holds for
{${\bf x}\in \mathbb{R}^n$}
and $\mathbf{v} = \mathbf{f} + \mathbbm{1}_{\{0 \}}(\mathbf{f}) \odot b(\xx)$.
\end{proposition}

\section{Evaluation} \label{sec:experiments}  

We conduct an experimental evaluation to answer the following questions. \\[-2em]
\begin{itemize}
\item[{\bf Q1}] Is CL-STE more scalable in injecting discrete constraints into neural network learning than existing neuro-symbolic learning methods? \\[-1em]
\item[{\bf Q2}] Does CL-STE make neural networks learn with no or fewer labeled data by effectively utilizing the given constraints?  \\[-1em]

\item[{\bf Q3}] Is CL-STE general enough to overlay constraint loss 
on different types of neural networks to enforce logical constraints and improve the accuracy of existing networks? 
\end{itemize}

Our implementation takes a CNF theory in DIMACS format (the standard format for input to SAT solvers).\footnote{All experiments in this section were done on Ubuntu 18.04.2 LTS with two 10-cores CPU Intel(R) Xeon(R) CPU E5-2640 v4 @ 2.40GHz and four GP104 [GeForce GTX 1080].}
Since the CL-STE method alone doesn't have associated symbolic rules,  unlike DeepProbLog, NeurASP, and NeuroLog, in this section, we compare these methods
 on the classification accuracy of the trained NNs (e.g., correctly predicting the label of an MNIST image) instead of query accuracy (e.g., correctly predicting the sum of two MNIST images). 

\NBB{CL-STE name has to be introduced} 

\subsection{mnistAdd Revisited}  \label{ssec:benchmarks}
We introduced the CNF encoding and the loss function for the {\bf mnistAdd} problem in Example~\ref{ex:mnistAdd}. The problem was used  in \cite{manhaeve18deepproblog} and \cite{yang20neurasp} as a benchmark.


\begin{figure}[ht!]
\begin{center}
\includegraphics[width=0.8\columnwidth]{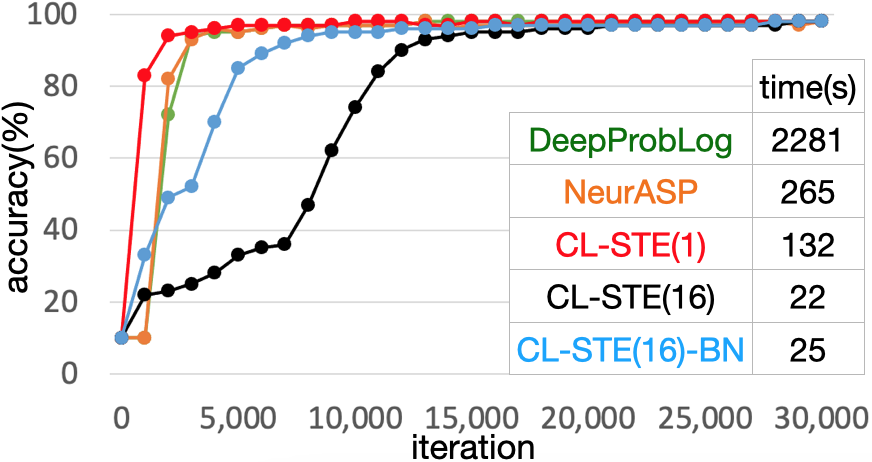}
\vspace{-4mm}
\caption{Comparison on mnistAdd}
\label{fig:mnistAdd}
\vspace{-3mm}
\end{center}

\end{figure}

Figure~\ref{fig:mnistAdd} compares the MNIST digit classification accuracy of neural networks trained by different methods on a single epoch of 30,000  addition data from \cite{manhaeve18deepproblog}.
``CL-STE($n$)" denotes our method with $b_p(x)$ and iSTE using a batch of size $n$.
%
As we see, DeepProbLog, NeurASP, and CL-STE with a batch size of 1 could quickly converge to near 100\% test accuracy.  Training time-wise, CL-STE outperforms the other approaches since it does not need to generate arithmetic circuits for every training instance as in DeepProbLog or enumerate all models as in NeurASP.
Also, while DeepProbLog and NeurASP do not support batch training, CL-STE could leverage the batch training to reduce the training time to 22s
with a batch size of 16 (denoted by CL-STE(16)). 
We observe that increasing the batch size in CL-STE also increases the number of parameter updates for convergence, which we could decrease by using batch normalization as shown in the blue line denoted by CL-STE(16)-BN.

Furthermore, we apply CL-STE to the variants of {\bf mnistAdd} by training with two-digit sums ({\bf mnistAdd2} \cite{manhaeve18deepproblog})  and three-digit sums ({\bf mnistAdd3}). 
Table~\ref{table:mnistAdd} shows that the CL-STE method scales much better than DeepProbLog and NeurASP. The time and accuracy are reported for a single epoch of training, where the cutoff time is 24 hours after which we report ``timeout.'' 

\vspace{-3mm}
\begin{table}
\centering
{\scriptsize
\caption{Experiments on {\bf mnistAdd}}
\label{table:mnistAdd}
\begin{tabular}{c| c | c | c} 
	\toprule
	& mnistAdd& mnistAdd2 & mnistAdd3  \\
	\hline
	DeepProbLog & 98.36\%  ~2565s & 97.57\% ~ 22699s & timeout  \\ \hline
	NeurASP & 97.87\%  ~~292s~~ & 97.85\% ~ 1682s~~ & timeout  \\ \hline
	CL-STE & 97.48\% ~~22s~~~~ & 98.12\% ~ 92s~~~~  & 97.78\% ~ 402s \\
	\bottomrule
\end{tabular}
}
	\vspace{-4mm}
\end{table}

 \subsection{Benchmarks from \cite{tsamoura21neural}} \label{ssec:weak-supervision}

The following are benchmark problems from \cite{tsamoura21neural}. 
Like the {\bf mnistAdd} problem, labels are not immediately associated with the data instances but with the results of logical operations applied to them. 

\noindent
{\bf add2x2}\ \ \  
The input is a $2 \times 2$ grid of digit images. The output is the four sums of the pairs of digits in each row/column. The task is to train a CNN for digit classification. 

\noindent
{\bf apply2x2}\ \ \ 
The input is three digits and a $2\times 2$ grid of hand-written math operator images in $\{+,-,\times \}$. The output is the four results of applying the two math operators in each row/column in the grid on the three digits. The task is to train a CNN for math operator classification.

\noindent
{\bf member(n)}\ \ \ 
The input is a set of $n$ images of digits and a digit in $\{0,\dots,9 \}$. The output is 0 or 1, indicating whether the digit appears in the set of digit images. The task is to train a CNN for digit classification.

Table~\ref{table:neurolog:domains} compares our method with DeepProbLog, NeurASP, and NeuroLog test accuracy-wise and training time-wise. 
Note that, instead of comparing the query accuracy as in \cite{tsamoura21neural}, we evaluate and compare the NN classification accuracies.

\begin{table}
\centering
{\scriptsize
\caption{Comparison between CL-STE and other approaches:
	{\small The numbers in parentheses are the times spent by NeuroLog to 
		generate all abductive proofs.}
}
\label{table:neurolog:domains}
\begin{tabular}{c|c c c c } 
	\toprule
	\multicolumn{1}{c}{}& add2x2 & apply2x2 & member(3) & member(5) \\
	\hline
	\multicolumn{1}{c}{accuracy(\%)}& \\
	\hline
	DeepProbLog & 88.4$\pm$0.7 & 100$\pm$0 & 96.3$\pm$0.3 & timeout \\ \hline
	NeurASP & 97.6$\pm$0.2 & 100$\pm$0 & 93.5$\pm$0.9 & timeout \\ \hline
	NeuroLog & 97.5$\pm$0.4 & 100$\pm$0 & 94.5$\pm$1.5 & 93.9$\pm$1.5 \\ \hline
	$b(x)$ + iSTE & 95.5$\pm$0.7 & 100$\pm$0 & 73.2$\pm$9.1 & 51.1$\pm$24.9 \\ 
	$b(x)$ + sSTE & 95.7$\pm$0.5 & 100$\pm$0 & 83.2$\pm$8.4 & 88.0$\pm$7.1 \\
	$b_p(x)$ + iSTE & 98.0$\pm$0.2 & 100$\pm$0 & 95.5$\pm$0.7 & 95.0$\pm$0.5 \\ 
	\hline
	\multicolumn{1}{c}{time(s)}& \\ 
	\hline
	DeepProbLog & 1035$\pm$71 & 586$\pm$9 & 2218$\pm$211 & timeout \\ \hline
	NeurASP & 142$\pm$2 & 47$\pm$1 & 253$\pm$1 & timeout \\ \hline
	\multirow{2}{*}{NeuroLog} & 2400$\pm$46 & 2428$\pm$29 & 427$\pm$12 & 682$\pm$40 \\
	& (1652) & (2266) & (27) & (114) \\ \hline
	$b(x)$ + iSTE & 80$\pm$2 & 208$\pm$1 & 45$\pm$0 & 177$\pm$1 \\ 
	$b(x)$ + sSTE & 81$\pm$2 & 214$\pm$8 & 46$\pm$1 & 181$\pm$10 \\
	$b_p(x)$ + iSTE & 54$\pm$4 & 112$\pm$2 & 43$\pm$3 & 49$\pm$4 \\ 
	\bottomrule
\end{tabular}
}
 	\vspace{-4mm}
\end{table}

Our experiments agree with \cite{yin19understanding},
which proves the instability issue of iSTE and the convergence guarantees with sSTE in a simple 2-layer CNN. Their experiments also observe a better performance of $b(x)$+sSTE over $b(x)$+iSTE on deep neural networks.
Our experimental results (especially for member(5) problem) also reveal the instability issue of $b(x)$+iSTE and show that $b(x)$+sSTE achieves higher and more stable accuracy.
Furthermore, we observe that $b_p(x)$ works better than $b(x)$ in terms of both accuracy and time in our experiments. This is because the input $x$ to $b_p(x)$ is normalized into probabilities before binarization, resulting in less information loss (i.e., change in magnitude ``$b_p(x)-x$'') when the neural network accuracy increases.

\subsection{CNN + Constraint Loss for Sudoku}  \label{ssec:sudoku}

The following experimental setting from \cite{yang20neurasp} demonstrates unsupervised learning with NeurASP on Sudoku problems. 
Given a {textual representation} of a Sudoku puzzle (in the form of a $9\times 9$ matrix where an empty cell is represented by 0), \citeauthor{park18can} (\citeyear{park18can}) trained a CNN (composed of 9 convolutional layers and a $1\times 1$ convoutional layer, followed by softmax) using 1 million examples and achieved 70\% test accuracy using an ``inference trick'': instead of predicting digits for all empty cells at once, which leads to poor accuracy, predict the most probable grid-cell value one by one. 
With the same CNN and inference trick, \citeauthor{yang20neurasp} (\citeyear{yang20neurasp}) achieved 66.5\% accuracy with only 7\% data with no supervision (i.e., 70k data instances without labels) by enforcing semantic constraints in neural network training with NeurASP. 
In this section, we consider the same unsupervised learning problem for Sudoku while we represent the Sudoku problem in CNF and use $L_{cnf}$ to enforce logical constraints during training. 

We use a CNF theory for $9\times 9$ Sudoku problems with $9^3=729$ atoms and $8991$ clauses as described in Appendix~\ref{appendix:subsec:cnf-sudoku}. This CNF can be represented by a matrix $\mathbf{C}\in\{-1,0,1 \}^{8991\times 729}$. 
{
The dataset consists of 70k data instances, 
80\%/20\% for training/testing.
Each data instance is a pair $\langle \mathbf{q},\mathbf{l} \rangle$ where $\mathbf{q}\in \{0,1,\dots,9 \}^{81}$ denotes a $9\times 9$ Sudoku board ($0$ denotes an empty cell) and $\mathbf{l}\in \{1,\dots,9 \}^{81}$ denotes its solution ($\mathbf{l}$ is not used in NeurASP and our method during training).
}
The non-zero values in $\mathbf{q}$ are treated as atomic facts $F$ and we construct the matrix $\mathbf{F} \in\{0,1\}^{81\times 9}$ such that, for $i\in \{ 1,\dots,81\}$, the $i$-th row $\mathbf{F}[i,:]$ is the vector $\{0\}^9$ if $\mathbf{q}[i]=0$ and is the one-hot vector for $\mathbf{q}[i]$ if $\mathbf{q}[i]\neq 0$. Then, the vector $\mathbf{f} \in\{0,1\}^{729}$ is simply the flattening of $\mathbf{F}$. We feed $\mathbf{q}$ into the CNN and obtain the output 
{
$\xx\in [0,1]^{729}$.
}
Finally, the prediction $\mathbf{v}\in\{0,1\}^{729}$ is obtained as $\mathbf{f} + \mathbbm{1}_{\{0 \}}(\mathbf{f}) \odot b_p(\xx)$, and the total loss function ${\cal L}$ we used is
{
$
{\cal L} =  L_{cnf}(\mathbf{C}, \mathbf{v}, \mathbf{f}) + 0.1 \times L_{bound}(\xx). 
$
}

\begin{table}
{\scriptsize 
\caption{CNN, NeurASP, and CL-STE on Park 70k Sudoku dataset (80\%/20\% split) w/ and w/o inference trick} 	
\label{tbl:sudoku:70k:new}
\begin{tabular}{c|c|c|c|r} \toprule
	Method & Supervised & $Acc_{wo}$ & $Acc_{w}$ & time(m)  \\ 
	\hline
	 Park's CNN & Full & 0.94\% & 23.3\% & 163 \\ 
	\hline
	Park's CNN+NeurASP & No & 1.69\% & 66.5\% & 13230  \\
	\hline
	Park's CNN+CL-STE & No & 2.38\% & 93.7\% & 813 \\ 
	\bottomrule   
\end{tabular}
} \\[-2em]
\end{table}

Table~\ref{tbl:sudoku:70k:new} compares the training time and the (whole-board) test accuracies with and without the inference trick ($Acc_w$ and $Acc_{wo}$, resp.) using $b_p(x)$+iSTE against NeurASP and baseline CNN \cite{park18can}. In each experiment, the same CNN is trained with only 70k (labeled/unlabeled) data instances
{from \cite{yang20neurasp}}
with an average of 43 given digits in a puzzle (min: 26, max: 77). 
As we can see, our method outperforms NeurASP in both accuracy and time. 
Accuracy-wise, the CNN model trained using CL-STE is 27.2\% more accurate than the CNN model trained using NeurASP  when we use the inference trick. 
%
Training time-wise, CL-STE is 16 times faster than NeurASP because we directly encode semantic constraints in a loss function, which saves the time to call a symbolic engine externally (e.g., {\sc clingo} to enumerate all stable models as in NeurASP). 

Table~\ref{tbl:sudoku:ste_vs_satnet1} compares CNN+CL-STE with SATNet trained on Park 70k and tested on both Park 70k and Palm Sudoku dataset \citep{palm18recurrent}.
While CNN is less tailored to logical reasoning than SATNet, our experiment shows that, when it is trained via CL-STE, it performs better than SATNet. 
\\[-2em]
\begin{table}[ht]
	\centering
	{\small
		\caption{SATNet vs. CNN+CL-STE}
		\label{tbl:sudoku:ste_vs_satnet1}
		\begin{tabular}{c|c|c|c|c} \toprule
			\multirow{2}*{Method} & Train Data & Test  & \multirow{2}*{\#Given} & 
			Test \\
			& (Supv) & Data &  & Accuracy    \\ 
			\hline
			\multirow{2}*{SATNet} & Park 70k & Park 70k & 26-77 & 67.78\%  \\ 
			\cline{3-5}
			& (Full) & Palm & 17-34 & 6.76\%   \\
			
			\hline
			\multirow{2}*{CNN+CL-STE} & Park 70k & Park 70k & 26-77 & 93.70\%  \\ 
			\cline{3-5}
			& (No) & Palm & 17-34 & 27.37\%   \\
			\bottomrule   
		\end{tabular}
	}
\end{table}

\vspace{-1em}

\subsection{GNN + Constraint Loss for Sudoku}

This section investigates if a GNN training can be improved with the constraint loss functions with STE by utilizing already known constraints without always relying on the labeled data. 
We consider the Recurrent Relational Network (RRN) \cite{palm18recurrent}, a state-of-the-art GNN for multi-step relational reasoning that achieves 96.6\% accuracy for hardest Sudoku problems by training on 180k labeled data instances. 
Our focus here is to make RRN learn more effectively using fewer data by injecting known constraints.


The training dataset in \cite{palm18recurrent} contains 180k data instances evenly distributed in 18 difficulties with 17-34 given numbers. We use a small subset of this dataset with random sampling.
Given a data instance $\langle \mathbf{q}, \mathbf{l} \rangle$ where $\mathbf{q}\in \{0,1,\dots,9 \}^{81}$ denotes a $9\times 9$ Sudoku board and $\mathbf{l}\in \{1,\dots,9 \}^{81}$ denotes its solution,
RRN takes $\mathbf{q}$ as input and, after 32 {iterations} of message passing, outputs 32 matrices of probabilities $\mathbf{X}_s \in \mathbf{R}^{81\times 9}$ where $s\in \{1,\dots,32\}$;  for example, $\mathbf{X}_1$ is the RRN prediction after 1 message passing step.

The baseline loss is the sum of the cross-entropy losses between prediction $\mathbf{X}_s$ and label $\mathbf{l}$ for all $s$. 

We evaluate if using constraint loss could further improve the performance of RRN with the same labeled data.
We use the same $L_{cnf}$ and $L_{bound}$ defined in CNN (with weights $1$ and $0.1$, resp.), which are applied to $\mathbf{X}_1$ only 
so that the RRN could be trained to deduce new digits in a single message passing step. 
We also use a continuous regularizer $L_{sum}$ below to regularize every $\mathbf{X}_s$ that ``the sum of the 9 probabilities in $\mathbf{X}_s$ in the same row/column/box must be 1'':
\begin{align*}
L_{sum} &= \sum\limits_{s\in \{1,\dots,32 \}\atop i\in \{row,col,box\}} avg((sum(\mathbf{X}_s^i) - 1)^2).
\end{align*}
Here, $avg(X)$ and $sum(X)$ compute the average and sum of all elements in $X$ along its last dimension;
$\mathbf{X}_s^{row}, \mathbf{X}_s^{col}, \mathbf{X}_s^{box} \in \mathbb{R}^{81\times 9}$ are reshaped copies of $\mathbf{X}_s$ such that each row in, for example, $\mathbf{X}_s^{row}$ contains 9 probabilities for atoms $a(1,C,N), \dots, a(9,C,N)$ for some $C$ and $N$. 

\begin{figure}
	\begin{center}
		\includegraphics[width=0.7\columnwidth]{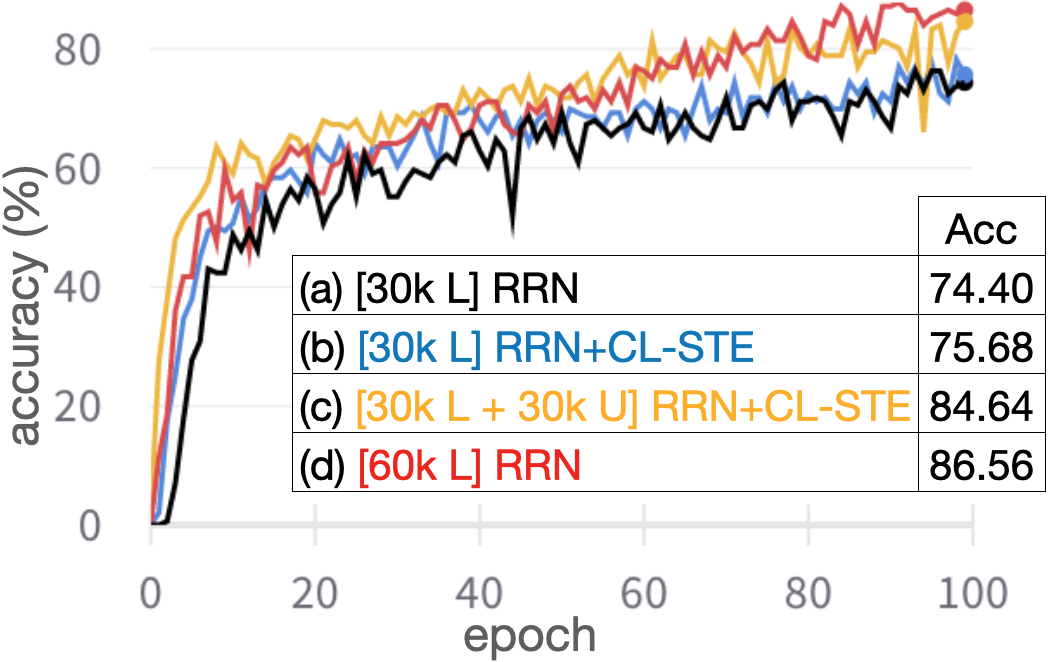}
		\vspace{-4mm}
				\caption{Test accuracy on the same randomly sampled 1k data from Palm Sudoku dataset when trained with RRN(+STE) with 30k to 60k [L]abeled/[U]nlabeled data} 
		\label{fig:rrn:cnf:30k}
  	\vspace{-3mm}
	\end{center}
\end{figure}

Figure~\ref{fig:rrn:cnf:30k} compares the test accuracy of the RRN trained for 100 epochs under 4 settings:
(a) the RRN trained with baseline loss using 30k labeled data;
(b) the RRN trained with both baseline loss and constraint losses
($L_{sum}$, $L_{cnf}$, and $L_{bound}$) using the same 30k labeled data; 
(c) the same setting as (b) with additional 30k unlabeled data;  
(d) same as (a) with additional 30k labeled data.
Comparing (a) and (b) indicates the effectiveness of the constraint loss using the same number of labeled data;  
comparison between (b) and (c) indicates even with the same number of labeled data but adding unlabeled data could increase the accuracy (due to the constraint loss); 
comparison between (c) and (d) shows that the effectiveness of the constraint loss is comparable to adding additional 30k labels. 

\begin{figure}
	\begin{center}
		\centerline{\includegraphics[width=0.85\columnwidth]{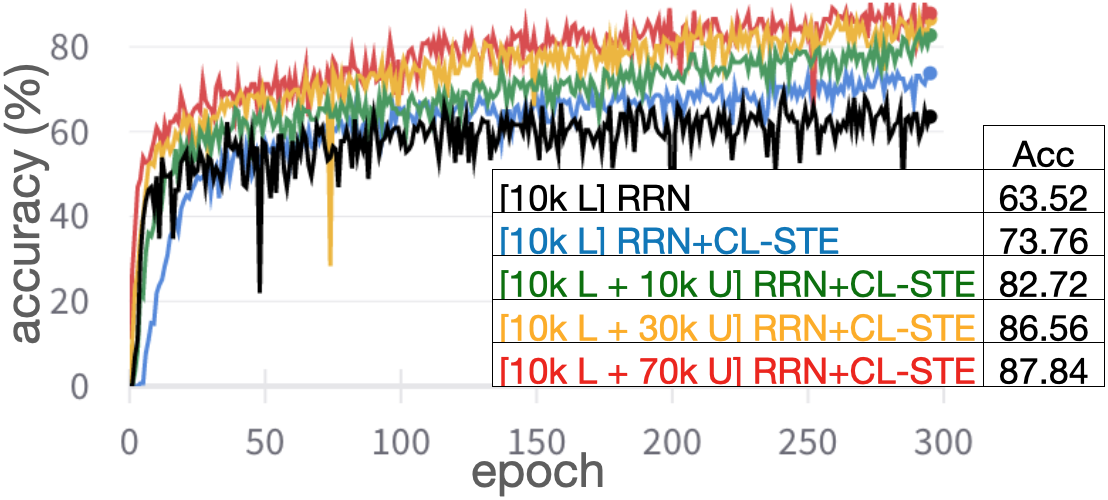}}
	\vspace{-4mm}
			\caption{Semi-supervised learning with RRN+STE on Sudoku using only 10k labeled data and varying numbers of unlabeled data from Palm dataset for training and using the same randomly sampled 1k data for testing}
		\label{fig:sudoku:rrn:semi:10k}
			\vspace{-9mm}
	\end{center}
\end{figure}

Figure~\ref{fig:sudoku:rrn:semi:10k} assesses the effect of constraint loss using fixed 10k labeled data and varying numbers (10k, 30k, 70k) of unlabeled data. 
We see that the baseline RRN trained with 10k labeled data ([10k L] RRN) has roughly saturated  while the other methods are still slowly improving the accuracy.
Training with the same number of labeled data but adding more unlabeled data makes the trained RRN achieve higher test accuracy, indicating that the constraint loss is effective in training even when the labels are unavailable.

\BOCC
{\bf Remark.}\ \ \ Although the CNN in the previous section seems to show comparable accuracies as the RRN in this section, 
{\cblu 
it relies on the inference trick and has to be trained on a dataset with varying difficulties (e.g., 26-77 givens) to make the inference trick work.
}
The RRN result does not use the trick, and we find that its learning is more robust.  On the other hand, the RRN could not learn well when there are no labeled data unlike the CNN. 
\EOCC

\subsection{Discussion} \label{subsec:discussio}
Regarding {\bf Q1}, Figure~\ref{fig:mnistAdd}, Tables~\ref{table:mnistAdd} and \ref{table:neurolog:domains} show that our method achieves comparable accuracy with existing neuro-symbolic formalisms but is much more scalable. 
Regarding {\bf Q2}, Table~\ref{tbl:sudoku:70k:new} and 
Figures~\ref{fig:rrn:cnf:30k} and \ref{fig:sudoku:rrn:semi:10k}
illustrate our method could be used for unsupervised and semi-supervised learning by utilizing the constraints underlying the data. 
Regarding {\bf Q3}, we applied constraint loss to MLP, CNN, and GNN, and observed that it improves the existing neural networks' prediction accuracy.

As we noted,  the gradient computation in other neuro-symbolic approaches, such as NeurASP, DeepProbLog, and NeuroLog, requires external calls to symbolic solvers to compute stable models or proofs for every data instance, which takes a long time. These approaches may give better quality gradients to navigate to feasible solutions, but their gradient computations are associated with NP-hardness (the worst case exponential size of SDD, computing all proofs or stable models).
In comparison, CL-STE treats each clause independently and locally to accumulate small pieces of gradients, allowing us to leverage GPUs and batch training as in the standard deep learning.
The method resembles local search and deduction in SAT, and the gradients may not reflect the global property but could be computed significantly faster.
Indeed, together with the gradient signals coming from the data, our method works well even when logical constraints are hard to satisfy, e.g., in training a neural network to solve Sudoku where a single feasible solution lies among $9^{47}$ to $9^{64}$ candidates when 17-34 digits are given.

\section{Conclusion} \label{sec:conclusion}



Constraint loss helps neural networks learn with fewer data, but the state-of-the-art methods require combinatorial computation to compute gradients. By leveraging STE, we demonstrate the feasibility of more scalable constraint learning in neural networks. Also, we showed that GNNs could learn with fewer (labeled) data by utilizing known constraints. Based on the formal properties of the CNF constraint loss and the promising initial experiments here, the next step is to apply the method to larger-scale experiments. 

\BOC
We could also consider an extension to weighted constraints. Although we treat all clauses as equal importance in learning, one can consider assigning weights to clauses so that more important constraints are prioritized in learning. Assigning weights to loss functions reflects the idea of weighted logics, such as Markov Logic \cite{richardson06markov}. 
\EOC

\section*{Acknowledgements}
\vspace{-0.5em}
We are grateful to Adam Ishay and the anonymous referees for their useful comments. This work was partially supported by the National Science Foundation under Grant IIS-2006747.
\vspace{-1em}


\bibliographystyle{icml2022}

\include{ste-ns-icml-camera-supp-0617}

\end{document}

%% file: ste-ns-icml-camera-supp-0617.tex
\appendix
~

{\Large Appendix }

The appendix contains the proofs of all propositions and more details about the experiments in the main body as well as additional experiments.

\section{CNF Loss with Basic Math Operations} \label{appendix:cnf_loss}
In Section~\ref{sec:ste-learning}, we defined the CNF loss using broadcasting, which is a common technique used for performing arithmetic operations between tensors having different shapes.\footnote{
\url{https://towardsdatascience.com/broadcasting-in-numpy-58856f926d73}
}
We could also give the definition of the CNF without referring to broadcasting as follows. 

Consider a propositional signature $\sigma = \{p_1, \dots, p_n\}$. 
Recall that we have
\begin{itemize}
\item the matrix $\mathbf{C} \in \{-1,0,1 \}^{m\times n}$ to represent the CNF theory such that $\mathbf{C}[i,j]$ is $1$ ($-1$, resp.) if $p_j$ ($\neg p_j$, resp.) belongs to the $i$-th clause in the theory, and is $0$ if neither $p_j$ nor $\neg p_j$ belongs to the clause;
\item the vector $\mathbf{f}\in \{0,1\}^{n}$ to represent $F$ such that $\mathbf{f}[j]$ is $1$ if $p_j\in F$ and is $0$ otherwise; and
\item the vector $\mathbf{v}\in \{0,1\}^{n}$ to represent $v$ such that $\mathbf{v}[j]$ is $1$ if $v(p_j)=\true$, and is $0$ if $v(p_j)=\false$.
\end{itemize}

Using $\mathbf{C}$, $\mathbf{v}$, and $\mathbf{f}$, 
we define the CNF loss $L_{cnf}(\mathbf{C}, \mathbf{v}, \mathbf{f})$ with basic math operations as follows where $i\in \{1,\dots,m\}$ and $j\in\{1,\dots,n \}$.
\begin{align}
\mathbf{L}_f[i,j] =\ & \mathbf{C}[i,j] \times \mathbf{f}[j]  \nonumber \\
\mathbf{L}_v[i,j] =\ & \mathbbm{1}_{\{1\}}(\mathbf{C}[i,j]) \times \mathbf{v}[j] ~~+ \nonumber \\ 
& \mathbbm{1}_{\{-1\}}(\mathbf{C}[i,j]) \times (1-\mathbf{v}[j])  \nonumber \\
\mathbf{deduce}[i] =\ & \mathbbm{1}_{\{1\}}\Big(\sum\limits_{j}
\big(|\mathbf{C}[i,j]|\big)
~~- \nonumber \\
& \hspace{8.5mm} \sum\limits_{j}\big(\mathbbm{1}_{\{-1\}}(\mathbf{L}_f[i,j])\big) \Big)  \nonumber \\
\mathbf{unsat}[i] =\ & \prod\limits_{j}\big(1-\mathbf{L}_v[i,j]\big) \nonumber \\
\mathbf{keep}[i] =\ & \sum\limits_{j}\Big(\mathbbm{1}_{\{1\}}(\mathbf{L}_v[i,j]) \times (1-\mathbf{L}_v[i,j]) ~+ \nonumber \\
& \hspace{8.5mm} \mathbbm{1}_{\{0\}}(\mathbf{L}_v[i,j]) \odot \mathbf{L}_v[i,j] \Big) \nonumber
\end{align}
\begin{align}
L_{deduce} =\ & \sum\limits_{i}\big(\mathbf{deduce}[i] \times \mathbf{unsat}[i]\big) \nonumber \\
L_{unsat} =\ & \frac{1}{n}\sum\limits_{i}\big( \mathbbm{1}_{\{1\}}(\mathbf{unsat}[i]) \times \mathbf{unsat}[i] \big)  \nonumber \\
L_{sat} =\ & \frac{1}{n}\sum\limits_{i}\big(\mathbbm{1}_{\{0\}}(\mathbf{unsat}[i]) \times \mathbf{keep}[i] \big) \nonumber 
\end{align}
{
\begin{align}
L_{cnf}(\mathbf{C}, \mathbf{v}, \mathbf{f}) =\ & L_{deduce} + L_{unsat} + L_{sat}.  \nonumber
\end{align}
}

\section{Proofs}
\noindent{\bf Proposition~\ref{prop:diode}~~}
When $K$ approaches $\infty$ and $|g(x)|\leq c$ for a constant $c$, the value of $\wt{b}^K(x)$ converges to $b(x)$.
\begin{align*}
\lim_{K \to \infty}\wt{b}^K(x) = b(x)
\end{align*}
The gradient $\frac{\partial \wt{b}^K(x)}{\partial x}$, whenever defined, is exactly the iSTE of $\frac{\partial b(x)}{\partial x}$ if $g(x)=1$, or the sSTE of $\frac{\partial b(x)}{\partial x}$ if 
\begin{align*}
g(x) = \begin{cases}
1 & \text{if $-1 \leq x \leq 1$}\\
0 & \text{otherwise.}
\end{cases}
\end{align*}
{\bf [Remark]}: Proposition~\ref{prop:diode} in our paper is similar to proposition 1 in \cite{kim20plug} but not the same. For the value of $\wt{b}^K(x)$, we don't have a condition that $g'(x)$ should be bounded. For the gradient of $\wt{b}^K(x)$, we have a stronger statement specific for STEs and don't have the condition for $K$ approaching $\infty$.

\begin{proof}	
Recall the definition of $\wt{b}^K(x)$
\begin{align*}
\wt{b}^K(x) = b(x) + s^K(x)g(x)
\end{align*}
where $K$ is a constant; $s^K(x) = \frac{Kx - \lfloor{Kx}\rfloor}{K}$ is a gradient tweaking function whose value is less than $\frac{1}{K}$ and whose gradient is always 1 whenever differentiable; and $g(x)$ is a gradient shaping function. 

\noindent{\bf [First]}, we will prove $\lim_{K \to \infty}\wt{b}^K(x) = b(x)$. Since $\wt{b}^K(x) = b(x) + s^K(x)g(x)$, it's equivalent to proving 
\begin{align*}
\lim_{K \to \infty} s^K(x)g(x) = 0
\end{align*}
Since $s^K(x) = \frac{Kx - \lfloor{Kx}\rfloor}{K}$ and $0\leq Kx - \lfloor{Kx}\rfloor \leq 1$, we know $0\leq s^K(x)\leq \frac{1}{K}$. Since $|g(x)|\leq c$ where $c$ is a constant, $-\frac{c}{K} \leq s^K(x)g(x) \leq \frac{c}{K}$. Thus $0 \leq \lim_{K \to \infty} s^K(x)g(x) \leq 0$, and consequently, $\lim_{K \to \infty} s^K(x)g(x) = 0$. 

\noindent{\bf [Second]}, we will prove
\begin{itemize}
\item when $g(x)=1$ and $s(x)=x$ (i.e., under iSTE), 
\begin{align*}
	\frac{\partial \wt{b}^K(x)}{\partial x} = 
	\begin{cases}
		\frac{\partial s(x)}{\partial x} & \text{(if $Kx\neq \lfloor{Kx}\rfloor$)}\\
		\text{undefined} & \text{otherwise.}
	\end{cases}
\end{align*}
\end{itemize}
Let's prove some general properties of the gradients.
Since $s(x)=x$, $g(x)=1$, and $s^K(x) = \frac{Kx - \lfloor{Kx}\rfloor}{K}$, 
\begin{itemize}
\item $\frac{\partial s(x)}{\partial x}=1$, $\frac{\partial g(x)}{\partial x}=0$, and
\item $\frac{\partial s^K(x)}{\partial x}=1$ whenever differentiable (i.e., whenever $Kx\neq \lfloor{Kx}\rfloor$).
\end{itemize}
Then, 
\begin{align*}
\frac{\partial \wt{b}^K(x)}{\partial x} &= 
\frac{\partial (b(x) + s^K(x)g(x))}{\partial x}\\
&= \frac{\partial (s^K(x)\times g(x))}{\partial x} \\
&= \frac{\partial s^K(x)}{\partial x}\\
&= 
\begin{cases}
	1 & \text{(if $Kx\neq \lfloor{Kx}\rfloor$)}\\
	\text{undefined} & \text{otherwise.}
\end{cases}
\end{align*}

\noindent{\bf [Third]}, we will prove
\begin{itemize}
\item when $g(x)=1$ if $-1\leq x\leq 1$ and $g(x)=0$ otherwise, and $s(x)=min(max(x, -1), 1)$ (i.e., under sSTE), 
\begin{align*}
	\frac{\partial \wt{b}^K(x)}{\partial x} = 
	\begin{cases}
		\frac{\partial s(x)}{\partial x} & \text{(if $Kx\neq \lfloor{Kx}\rfloor$)}\\
		\text{undefined} & \text{otherwise.}
	\end{cases}
\end{align*}
\end{itemize}
Let's prove some general properties of the gradients.
Since $s(x)=min(max(x, -1), 1)$, $g(x)=1$ if $-1\leq x\leq 1$ and $g(x)=0$ otherwise, and $s^K(x) = \frac{Kx - \lfloor{Kx}\rfloor}{K}$, 
\begin{itemize}
\item $\frac{\partial s(x)}{\partial x}=1$ if $-1\leq x\leq 1$ and $\frac{\partial s(x)}{\partial x}=0$ otherwise,
\item $\frac{\partial g(x)}{\partial x}=0$, and
\item $\frac{\partial s^K(x)}{\partial x}=1$ whenever differentiable (i.e., whenever $Kx\neq \lfloor{Kx}\rfloor$).
\end{itemize}
Then, 
\begin{align*}
\frac{\partial \wt{b}^K(x)}{\partial x} &= 
\frac{\partial (b(x) + s^K(x)g(x))}{\partial x}\\
&= \frac{\partial (s^K(x)\times g(x))}{\partial x} \\
&= g(x) \times \frac{\partial s^K(x)}{\partial x} + s^K(x) \times \frac{\partial g(x)}{\partial x}\\
&= g(x) \times \frac{\partial s^K(x)}{\partial x}\\
&= 
\begin{cases}
	g(x) & \text{(if $Kx\neq \lfloor{Kx}\rfloor$)}\\
	\text{undefined} & \text{otherwise.}
\end{cases}\\
&= 
\begin{cases}
	\frac{\partial s(x)}{\partial x} & \text{(if $Kx\neq \lfloor{Kx}\rfloor$)}\\
	\text{undefined} & \text{otherwise.}
\end{cases}
\end{align*}
\end{proof} 

\bigskip
\noindent{\bf Proposition~\ref{prop:value}~~}
Given a CNF theory $C$, a set $F$ of atoms, and a truth assignment $v$ such that $v\models F$, let $\mathbf{C}, \mathbf{f}, \mathbf{v}$ denote their matrix/vector representations, respectively. Let $C_{deduce} \subseteq C$ denote the set of Horn clauses $H$ in $C$ such that all but one literal in $H$ are of the form $\neg p$ where $p\in F$.
Then 
\begin{itemize}
\item 
the minimum values of $L_{deduce}$, $L_{unsat}$, $L_{sat}$, and $L_{cnf}(\mathbf{C}, \mathbf{v}, \mathbf{f})$ are 0;
\item $v\models C_{deduce}$ iff $L_{deduce}$ is 0;
\item $v\models C$ iff $L_{unsat}$ is 0 iff $L_{cnf}(\mathbf{C}, \mathbf{v}, \mathbf{f})$ is 0.
\end{itemize}

\begin{proof}
Recall the definition of $L_{cnf}$
\begin{align*}
\mathbf{L}_f =\ & \mathbf{C} \odot \mathbf{f}  \\
\mathbf{L}_v =\ & \mathbbm{1}_{\{1\}}(\mathbf{C}) \odot \mathbf{v} + \mathbbm{1}_{\{-1\}}(\mathbf{C}) \odot (1-\mathbf{v}) \\
\mathbf{deduce} =\ & \mathbbm{1}_{\{1\}}\Big(sum(\mathbf{C} \odot \mathbf{C}) - sum(\mathbbm{1}_{\{-1\}}(\mathbf{L}_f)) \Big)  \\
\mathbf{unsat} =\ & prod(1-\mathbf{L}_v)  \\
\mathbf{keep} =\ & sum(\mathbbm{1}_{\{1\}}(\mathbf{L}_v) \odot (1-\mathbf{L}_v) + \mathbbm{1}_{\{0\}}(\mathbf{L}_v) \odot \mathbf{L}_v) \\
L_{deduce} =\ & sum(\mathbf{deduce} \odot \mathbf{unsat}) \\
L_{unsat} =\ & avg( \mathbbm{1}_{\{1\}}(\mathbf{unsat}) \odot \mathbf{unsat} )  \\
L_{sat} =\ & avg(\mathbbm{1}_{\{0\}}(\mathbf{unsat}) \odot \mathbf{keep} )
\end{align*}
\begin{align*}
L_{cnf}(\mathbf{C}, \mathbf{v}, \mathbf{f}) =\ & L_{deduce} + L_{unsat} + L_{sat}
\end{align*}
and the definitions of $\mathbf{C}, \mathbf{f}, \mathbf{v}$ below.
\begin{itemize}
\item the matrix $\mathbf{C}$ is in $\{-1,0,1 \}^{m\times n}$ such that $\mathbf{C}[i,j]$ is $1$ ($-1$, resp.) if $p_j$ ($\neg p_j$, resp.) belongs to the $i$-th clause, and is $0$ if neither $p_j$ nor $\neg p_j$ belongs to the clause; 
\item the vector $\mathbf{f}$ is in $ \{0,1\}^{n}$ to represent $F$ such that $\mathbf{f}[j]$ is $1$ if $p_j\in F$ and is $0$ otherwise; and
\item the vector $\mathbf{v}$ is in $ \{0,1\}^{n}$ to represent $v$ such that $\mathbf{v}[j]$ is $1$ if $v(p_j)=\true$, and is $0$ if $v(p_j)=\false$.
\end{itemize}
We will prove each bullet in Proposition~\ref{prop:value} as follows.

\begin{enumerate}
\item
{\bf [First]}, we will prove
\begin{itemize}
	\item $\mathbf{L}_f$ is the matrix in $\{-1,0,1\}^{m\times n}$ such that (i) $\mathbf{L}_f[i,j]=1$ iff clause $i$ contains literal $p_j$ and $p_j\in F$; and (ii) $\mathbf{L}_f[i,j]=-1$ iff clause $i$ contains literal $\neg p_j$ and $p_j\in F$.
	\item $\mathbf{L}_v$ is the matrix in $\{0,1\}^{m\times n}$ such that $\mathbf{L}_v[i,j]=1$ iff clause $i$ contains a literal ($p_j$ or $\neg p_j$) for atom $p_j$ and this literal evaluates to $\true$ under $v$.
\end{itemize}
According to the definition, $\mathbf{L}_f[i,j]=\mathbf{C}[i,j] \times \mathbf{f}[j]$. Since $\mathbf{f}[j] \in \{0,1\}$, we have $\mathbf{L}_f[i,j]=1$ iff ``$\mathbf{C}[i,j]=1$ and $\mathbf{f}[j]=1$'', and according to the definition of $\mathbf{C}$ and $\mathbf{f}$, we have $\mathbf{L}_f[i,j]=1$ iff ``clause $i$ contains literal $p_j$ and $p_j\in F$''.
Similarly, we have $\mathbf{L}_f[i,j]=-1$ iff ``$\mathbf{C}[i,j]=-1$ and $\mathbf{f}[j]=1$'' iff ``clause $i$ contains literal $\neg p_j$ and $p_j\in F$''.

According to the definition, $\mathbf{L}_v[i,j]=\mathbbm{1}_{\{1\}}(\mathbf{C})[i,j] \times \mathbf{v}[j] + \mathbbm{1}_{\{-1\}}(\mathbf{C})[i,j] \times (1-\mathbf{v}[j])$. 
Since $\mathbbm{1}_{\{1\}}(\mathbf{C})[i,j]$ and $\mathbbm{1}_{\{-1\}}(\mathbf{C})[i,j]$ cannot be $1$ at the same time and $\mathbf{v}[j] \in \{0,1\}$, we have $\mathbf{L}_v[i,j]=1$ iff ``$\mathbf{C}[i,j]=1$ and $\mathbf{v}[j]=1$'' or ``$\mathbf{C}[i,j]=-1$ and $\mathbf{v}[j]=0$''. According to the definition of $\mathbf{C}$ and $\mathbf{v}$, we have $\mathbf{L}_v[i,j]=1$ iff ``clause $i$ contains literal $p_j$, which evaluates to $\true$ under $v$'' or ``clause $i$ contains literal $\neg p_j$, which evaluates to $\true$ under $v$''.

{\bf [Second]}, we will prove
\begin{itemize}
	\item 
	$\mathbf{deduce}$ is the vector in $\{0,1\}^{m}$ such that $\mathbf{deduce}[i]=1$ iff clause $i$ has all but one literal of the form $\neg p_j$ such that $p_j \in F$.
	\item $\mathbf{unsat}$ is the vector in $\{0,1\}^{m}$ such that $\mathbf{unsat}[i]=1$ iff clause $i$ evaluates to $\false$ under $v$.
	\item $\mathbf{keep}$ is the vector $\{0\}^{m}$.
\end{itemize}

From the definition of $\mathbf{C}$, the matrix $\mathbf{C} \odot \mathbf{C}$ is in $\{0,1\}^{m\times n}$ such that the element at position ($i$,$j$) is 1 iff clause $i$ contains a literal ($p_j$ or $\neg p_j$) for atom $p_j$.
Since $sum(X)$ computes the sum of elements in each row of matrix $X$, for $i\in\{1,\dots,m\}$ and $k\in \{1,\dots,n\}$, $sum(\mathbf{C} \odot \mathbf{C})[i]=k$ iff clause $i$ contains $k$ literals. 
Recall that we proved that $\mathbf{L}_f[i,j]=-1$ iff ``clause $i$ contains literal $\neg p_j$ and $p_j\in F$''. 
Consequently, $\mathbbm{1}_{\{-1\}}(\mathbf{L}_f)$ is the matrix in $\{0,1\}^{m\times n}$ such that $\mathbbm{1}_{\{-1\}}(\mathbf{L}_f)[i,j]=1$ iff ``clause $i$ contains literal $\neg p_j$ and $p_j\in F$''.
As a result, $sum(\mathbf{C} \odot \mathbf{C}) - sum(\mathbbm{1}_{\{-1\}}(\mathbf{L}_f))$ is the vector in $\{0,\dots,n\}^m$ such that its $i$-th element is 1 iff ``clause $i$ contains all but one literal of the form $\neg p_j$ such that $p_j \in F$''. 
Thus $\mathbf{deduce}$ is the vector in $\{0,1\}^{m}$ such that $\mathbf{deduce}[i]=1$ iff ``clause $i$ has all but one literal of the form $\neg p_j$ such that $p_j \in F$''.

Since $prod(X)$ computes the product of elements in each row of matrix $X$, for $i\in \{1,\dots,m\}$, $\mathbf{unsat}[i] = \prod\limits_{j\in\{1,\dots,n\}} (1-\mathbf{L}_v[i,j])$.
Recall that we proved that
$\mathbf{L}_v$ is the matrix in $\{0,1\}^{m\times n}$ such that $\mathbf{L}_v[i,j]=1$ iff clause $i$ contains a literal ($p_j$ or $\neg p_j$) for atom $p_j$ and this literal evaluates to $\true$ under $v$.
Thus $\mathbf{unsat}[i]\in \{0,1\}$ and $\mathbf{unsat}[i]=1$ iff ``$\mathbf{L}_v[i,j]=0$ for $j\in \{1,\dots,n\}$'' iff ``for $j\in \{1,\dots,n\}$, clause $i$ either does not contain a literal for atom $p_j$ or contains a literal for atom $p_j$ while this literal evaluates to $\false$ under $v$'' iff ``clause $i$ evaluates to $\false$ under $v$''. 
In other words, $\mathbf{unsat}$ is the vector in $\{0,1\}^{m}$ such that $\mathbf{unsat}[i]=1$ iff clause $i$ evaluates to $\false$ under $v$.

Since $\mathbf{L}_v$ is the matrix in $\{0,1\}^{m\times n}$, for $i\in\{1,\dots,m\}$ and $j\in\{1,\dots,n\}$, $\mathbbm{1}_{\{1\}}(\mathbf{L}_v)[i,j]=1$ iff $\mathbf{L}_v[i,j]=1$ iff $(1-\mathbf{L}_v[i,j])=0$. Thus $\mathbbm{1}_{\{1\}}(\mathbf{L}_v) \odot (1-\mathbf{L}_v)$ is the matrix $\{0\}^{m\times n}$ of all zeros. Similarly, $\mathbbm{1}_{\{0\}}(\mathbf{L}_v) \odot \mathbf{L}_v)$ is also the matrix $\{0\}^{m\times n}$. As a result, $\mathbf{keep}$ is the vector $\{0\}^{m}$.

{\bf [Third]}, we will prove
\begin{itemize}
	\item $L_{deduce}$ is an integer in $\{0,\dots,m\}$ such that $L_{deduce}=k$ iff there are $k$ clauses in $C_{deduce}$ that are evaluated to $\false$ under $v$.
	\item $L_{unsat}$ is a number in $\{0, \frac{1}{m},\dots,\frac{m}{m}\}$ such that $L_{unsat}=\frac{k}{m}$ iff there are $k$ clauses that are evaluated to $\false$ under $v$.
	\item $L_{sat}$ is $0$.
\end{itemize}
Recall that we proved that $\mathbf{deduce}$ is the vector in $\{0,1\}^{m}$ such that $\mathbf{deduce}[i]=1$ iff clause $i$ has all but one literal of the form $\neg p_j$ such that $p_j \in F$; and $\mathbf{unsat}$ is the vector in $\{0,1\}^{m}$ such that $\mathbf{unsat}[i]=1$ iff clause $i$ evaluates to $\false$ under $v$.
According to the definition of $C_{deduce}$, $\mathbf{deduce}\odot \mathbf{unsat}$ is the vector in $\{0,1\}^{m}$ such that its $i$-th element is $1$ iff clause $i$ is in $C_{deduce}$ and evaluates to $\false$ under $v$. As a result, $L_{deduce}$ is an integer in $\{0,\dots,m\}$ such that $L_{deduce}=k$ iff there are $k$ clauses in $C_{deduce}$ that are evaluated as $\false$ under $v$.

Since $\mathbf{unsat}$ is the vector in $\{0,1\}^{m}$ such that $\mathbf{unsat}[i]=1$ iff clause $i$ evaluates to $\false$ under $v$, and since $\mathbf{unsat}[i]=1$ iff $\mathbbm{1}_{\{1\}}(\mathbf{unsat})[i]=1$, we know the $i$-th element in $\mathbbm{1}_{\{1\}}(\mathbf{unsat}) \odot \mathbf{unsat}$ is $1$ iff clause $i$ evaluates to $\false$ under $v$.  $L_{unsat}=avg(\mathbbm{1}_{\{1\}}(\mathbf{unsat}) \odot \mathbf{unsat})$ is a number in $\{0, \frac{1}{m},\dots,\frac{m}{m}\}$ such that $L_{unsat}=\frac{k}{m}$ iff there are $k$ clauses that are evaluated as $\false$ under $v$.

Recall that we proved that $\mathbf{keep}$ is the vector $\{0\}^{m}$. Thus $\mathbbm{1}_{\{0\}}(\mathbf{unsat}) \odot \mathbf{keep}$ is the vector $\{0\}^{m}$. Thus $L_{sat}$ is 0.

{\bf [Fourth]}, we will prove
\begin{itemize}
	\item the minimum values
	of $L_{deduce}$, $L_{unsat}$, $L_{sat}$, $L_{cnf}(\mathbf{C}, \mathbf{v}, \mathbf{f})$ are 0.
\end{itemize}
Recall that we proved that 
$L_{deduce}$ is an integer in $\{0,\dots,m\}$,
$L_{unsat}$ is a number in $\{0, \frac{1}{m},\dots,\frac{m}{m}\}$, and
$L_{sat}$ is 0. 
It's obvious that the minimum values of $L_{deduce}$, $L_{unsat}$, and $L_{sat}$ are 0.
Since (i) $L_{cnf}(\mathbf{C}, \mathbf{v}, \mathbf{f}) = L_{deduce} + L_{unsat} + L_{sat}$, (ii) $L_{deduce}=0$ when all clauses in $C_{deduce}$ are evaluated to $\true$ under $v$, and (iii) $L_{unsat}=0$ when all clauses in $C$ are evaluated to $\true$ under $v$, the minimum value of $L_{cnf}(\mathbf{C}, \mathbf{v}, \mathbf{f})$ is 0 and is achieved when all clauses in $C$ are evaluated to $\true$ under $v$.

\item We will prove
\begin{itemize}
	\item $v\models C_{deduce}$ iff $L_{deduce}$ is 0.
\end{itemize}
Recall that we proved that $L_{deduce}$ is an integer in $\{0,\dots,m\}$ such that $L_{deduce}=k$ iff there are $k$ clauses in $C_{deduce}$ that are evaluated as $\false$ under $v$. Then $L_{deduce}$ is 0 iff ``there is no clause in $C_{deduce}$ that evaluates to $\false$ under $v$'' iff ``every clause in $C_{deduce}$ evaluates to $\true$ under $v$'' iff $v\models C_{deduce}$.

\item We will prove
\begin{itemize}
	\item $v\models C$ iff $L_{unsat}$ is 0 iff $L_{cnf}(\mathbf{C}, \mathbf{v}, \mathbf{f})$ is 0.
\end{itemize}
Recall that we proved that $L_{unsat}$ is a number in $\{0, \frac{1}{m},\dots,\frac{m}{m}\}$ such that $L_{unsat}=\frac{k}{m}$ iff there are $k$ clauses that are evaluated as $\false$ under $v$. Then $L_{unsat}$ is 0 iff ``there is no clause in $C$ that evaluates to $\false$ under $v$'' iff $v\models C$.

Assume $L_{unsat}$ is 0. Then ``there is no clause in $C$ that evaluates to $\false$ under $v$''. Consequently, ``there is no clause in $C_{deduce}$ that is evaluated to $\false$ under $v$''. Recall that we proved that $L_{deduce}$ is an integer in $\{0,\dots,m\}$ such that $L_{deduce}=k$ iff there are $k$ clauses in $C_{deduce}$ that are evaluated as $\false$ under $v$. Then $L_{deduce}$ is 0. Since $L_{sat}$ is 0, $L_{cnf}(\mathbf{C}, \mathbf{v}, \mathbf{f})$ is 0.

Assume $L_{cnf}(\mathbf{C}, \mathbf{v}, \mathbf{f})$ is 0, which is the minimum value $L_{cnf}$ can take. It is easy to see that $L_{unsat}$ must be $0$.
\end{enumerate}
\end{proof} 

\bigskip

\noindent{\bf Proposition~\ref{prop:gradient:b_p}~~} 
Given a CNF theory $C$ of $m$ clauses and $n$ atoms and a set $F$ of atoms such that $C\cup F$ is satisfiable, let $\mathbf{C}, \mathbf{f}$ denote their matrix/vector representations, respectively. 
Given a neural network output {$\mathbf{x}\in [0,1]^{n}$ denoting probabilities},
we construct $\mathbf{v} = \mathbf{f} + \mathbbm{1}_{\{0 \}}(\mathbf{f}) \odot b_p(\xx)$ 
and a truth assignment $v$ such that $v(p_j)=\true$ if $\mathbf{v}[j]$ is $1$, and $v(p_j)=\false$ if $\mathbf{v}[j]$ is $0$. 
Let $C_{deduce} \subseteq C$ denote the set of Horn clauses $H$ in $C$ such that all but one literal in $H$ are of the form $\neg p$ and $p\in F$.
Then, for any $j\in \{1,\dots,n\}$,
\begin{enumerate}
\item if $p_j\in F$, all of $\frac{\partial L_{deduce}}{\partial \xx[j]}$, $\frac{\partial L_{unsat}}{\partial \xx[j]}$, and $\frac{\partial L_{sat}}{\partial \xx[j]}$ are zeros;
\item if $p_j\not \in F$,
\begin{align*}
\frac{\partial L_{deduce}}{\partial \xx[j]}& \stackrel{iSTE}{\approx}
\begin{cases}
	-c & \text{if $c>0$ clauses in $C_{deduce}$} \\
	&\text{contain literal $p_j$;}\\
	c & \text{if $c>0$ clauses in $C_{deduce}$}\\
	&\text{contain literal $\neg p_j$;} \\
	0 & \text{otherwise;}
\end{cases} \\
\frac{\partial L_{unsat}}{\partial \xx[j]} &
\stackrel{iSTE}{\approx}
\frac{c_2 - c_1}{m} \\
\frac{\partial L_{sat}}{\partial \xx[j]} &
\stackrel{iSTE}{\approx}
\begin{cases}
	-\frac{c_3}{m} & \text{if $v\models p_j$,} \\
	\frac{c_3}{m} & \text{if $v\not \models p_j$.} 
\end{cases}
\end{align*}
where $\stackrel{iSTE}{\approx}$ stands for the equivalence of gradients under iSTE; $c_1$ (and $c_2$, resp.) is the number of clauses in $C$ that are not satisfied by $v$ and contain $p_j$ (and $\neg p_j$, resp.); $c_3$ is the number of clauses in $C$ that are satisfied by $v$ and contain $p_j$ or $\neg p_j$.
\end{enumerate}

\begin{proof}
We will prove each bullet in Proposition~\ref{prop:gradient:b_p} as follows.
\begin{enumerate}
\item 
Take any $k\in\{1,\dots,n\}$, let's focus on $\xx[k]$ and compute the gradient of $L\in \{L_{deduce},L_{unsat},L_{sat} \}$ to it with iSTE. According to the chain rule and since $\frac{\partial \mathbf{v}[i]}{\partial b_p(\xx)[j]}=0$ for $i\neq j$, we have
\begin{align*}
	\frac{\partial L}{\partial \xx[k]}
	=& \frac{\partial L}{\partial \mathbf{v}[k]} 
	\times \frac{\partial \mathbf{v}[k]}{\partial b_p(\xx[k])}
	\times \frac{\partial b_p(\xx[k])}{\partial \xx[k]}.
\end{align*}
Under iSTE, the last term $\frac{\partial b_p(\xx[k])}{\partial \xx[k]}$ is replaced with $\frac{\partial s(\xx[k])}{\partial \xx[k]} = \frac{\partial \xx[k]}{\partial \xx[k]}=1$. Thus
\begin{align*}
	\frac{\partial L}{\partial \xx[k]}
	=& \frac{\partial L}{\partial \mathbf{v}[k]} 
	\times \frac{\partial \mathbf{v}[k]}{\partial b_p(\xx[k])}
	\times \frac{\partial b_p(\xx[k])}{\partial \xx[k]}\\
	\stackrel{iSTE}{\approx} & \frac{\partial L}{\partial \mathbf{v}[k]} 
	\times \frac{\partial \mathbf{v}[k]}{\partial b_p(\xx[k])} ~~\text{(under iSTE)}\\
	=& \frac{\partial L}{\partial \mathbf{v}[k]} 
	\times \frac{\partial (\mathbf{f}[k] + \mathbbm{1}_{\{0 \}}(\mathbf{f}[k]) \times b_p(\xx[k]))}{\partial b_p(\xx[k])}\\
	=&
	\begin{cases}
		\frac{\partial L}{\partial \mathbf{v}[k]} & \text{if $\mathbf{f}[k]=0$,}\\
		0 & \text{if $\mathbf{f}[k]=1$.}
	\end{cases}
\end{align*}
Since $\mathbf{f}[k]=1$ iff $p_k\in F$, if $p_k\in F$, then all of $\frac{\partial L_{deduce}}{\partial \xx[k]}$, $\frac{\partial L_{unsat}}{\partial \xx[k]}$, and $\frac{\partial L_{sat}}{\partial \xx[k]}$ are zeros.
\item 
Recall the definition of $L_{cnf}$
\begin{align*}
	\mathbf{L}_f =\ & \mathbf{C} \odot \mathbf{f}  \\
	\mathbf{L}_v =\ & \mathbbm{1}_{\{1\}}(\mathbf{C}) \odot \mathbf{v} + \mathbbm{1}_{\{-1\}}(\mathbf{C}) \odot (1-\mathbf{v}) \\
	\mathbf{deduce} =\ & \mathbbm{1}_{\{1\}}\Big(sum(\mathbf{C} \odot \mathbf{C}) - sum(\mathbbm{1}_{\{-1\}}(\mathbf{L}_f)) \Big)  \\
	\mathbf{unsat} =\ & prod(1-\mathbf{L}_v)  \\
	\mathbf{keep} =\ & sum(\mathbbm{1}_{\{1\}}(\mathbf{L}_v) \odot (1-\mathbf{L}_v) + \mathbbm{1}_{\{0\}}(\mathbf{L}_v) \odot \mathbf{L}_v) \\
	L_{deduce} =\ & sum(\mathbf{deduce} \odot \mathbf{unsat}) \\
	L_{unsat} =\ & avg( \mathbbm{1}_{\{1\}}(\mathbf{unsat}) \odot \mathbf{unsat} )  \\
	L_{sat} =\ & avg(\mathbbm{1}_{\{0\}}(\mathbf{unsat}) \odot \mathbf{keep} )
\end{align*}
\begin{align*}
	L_{cnf}(\mathbf{C}, \mathbf{v}, \mathbf{f}) =\ & L_{deduce} + L_{unsat} + L_{sat}
\end{align*}
We know $p_k\not\in F$ iff $\mathbf{f}[k]=0$. As proved in the first bullet, for $L\in \{L_{deduce},L_{unsat},L_{sat} \}$, if $p_k\not\in F$, then $\frac{\partial L}{\partial \xx[k]} \stackrel{iSTE}{\approx} \frac{\partial L}{\partial \mathbf{v}[k]}$. We further analyze the value of $\frac{\partial L}{\partial \mathbf{v}[k]}$ for each $L$ under the condition that $\mathbf{f}[k]=0$.

\medskip
[$L_{deduce}$]~~According to the definition,
{\small 
	\begin{align*}
		&L_{deduce} \\
		=& \sum\limits_{i\in \{1,\dots,m\}}\Big( \mathbf{deduce}[i] \times \mathbf{unsat}[i] \Big)\\
		=& \sum\limits_{i\in \{1,\dots,m\}}\Big( \mathbf{deduce}[i] \times \prod\limits_{j\in\{1,\dots,n\}}(1-\mathbf{L}_v[i,j]) \Big)
	\end{align*}
	\begin{align*}
		&\frac{\partial L_{deduce}}{\partial \mathbf{v}[k]} \\
		=& \sum\limits_{i\in \{1,\dots,m\}}\frac{\partial \Big( \mathbf{deduce}[i] \times \prod\limits_{j\in\{1,\dots,n\}}(1-\mathbf{L}_v[i,j])\Big)}{\partial \mathbf{v}[k]}\\
		=& \sum\limits_{i\in \{1,\dots,m\}}\Big(
		\frac{\partial \mathbf{deduce}[i]}{\partial \mathbf{v}[k]} \times \prod\limits_{j\in\{1,\dots,n\}}(1-\mathbf{L}_v[i,j]) +\\
		& \hspace{1.8cm} \mathbf{deduce}[i] \times
		\frac{\partial \prod\limits_{j\in\{1,\dots,n\}}(1-\mathbf{L}_v[i,j])}{\partial \mathbf{v}[k]} \Big)
	\end{align*}
}%
Since $\mathbf{deduce}$ is the result of an indicator function, $\frac{\partial \mathbf{deduce}[i]}{\partial \mathbf{v}[k]}=0$. Then,
{\small
	\begin{align*}
		&\frac{\partial L_{deduce}}{\partial \mathbf{v}[k]} \\
		=& \sum\limits_{i\in \{1,\dots,m\}}\Big(\mathbf{deduce}[i] \times
		\frac{\partial \prod\limits_{j\in\{1,\dots,n\}}(1-\mathbf{L}_v[i,j])}{\partial \mathbf{v}[k]} \Big).
	\end{align*}
}%
Let $U\subseteq \{1,\dots,m\}$ denote the set of indices of all clauses in $C_{deduce}$. Since $\mathbf{deduce}[i]=1$ iff $i\in U$,
\begin{align*}
	\frac{\partial L_{deduce}}{\partial \mathbf{v}[k]} =& \sum\limits_{i\in U}\Big(\frac{\partial \prod\limits_{j\in\{1,\dots,n\}}(1-\mathbf{L}_v[i,j])}{\partial \mathbf{v}[k]} \Big).
\end{align*}
Let $G_{i,k}$ denote $\frac{\partial \prod\limits_{j\in\{1,\dots,n\}}(1-\mathbf{L}_v[i,j])}{\partial \mathbf{v}[k]}$. Then
\begin{align*}
	\frac{\partial L_{deduce}}{\partial \mathbf{v}[k]} =& \sum\limits_{i\in U}G_{i,k}.
\end{align*}
Let's analyze the value of $G_{i,k}$ where $i\in U$ and $k\in \{1,\dots,n\}$ such that $\mathbf{f}[k]=0$.
According to the product rule below, 
\begin{align*}
	\frac{d}{d x}\left[\prod_{i=1}^{k} f_{i}(x)\right]=\left(\prod_{i=1}^{k} f_{i}(x)\right)\left(\sum_{i=1}^{k} \frac{f_{i}^{\prime}(x)}{f_{i}(x)}\right)
\end{align*}
we have
{\small
	\begin{align*}
		&G_{i,k}\\
		=& \frac{\partial \prod\limits_{j\in\{1,\dots,n\}}(1-\mathbf{L}_v[i,j])}{\partial \mathbf{v}[k]} \\
		=& \left(\prod\limits_{j\in\{1,\dots,n\}}(1-\mathbf{L}_v[i,j])\right)
		\times \sum\limits_{j\in\{1,\dots,n\}}
		\frac{\frac{\partial (1-\mathbf{L}_v[i,j])}{\partial \mathbf{v}[k]}}{1-\mathbf{L}_v[i,j]}
	\end{align*}
}%
Since $\mathbf{L}_v[i,j]=\mathbbm{1}_{\{1\}}(\mathbf{C})[i,j] \times \mathbf{v}[j] + \mathbbm{1}_{\{-1\}}(\mathbf{C})[i,j] \times (1-\mathbf{v}[j])$, we know
\begin{itemize}
	\item [{\bf (a)}] for $j\in\{1,\dots,n\}$ such that $j\neq k$, $\frac{\partial (1-\mathbf{L}_v[i,j])}{\partial \mathbf{v}[k]} = 0$ and $\frac{\partial \mathbf{L}_v[i,j]}{\partial \mathbf{v}[k]} = 0$;
	\item [{\bf (b)}] when clause $i$ doesn't contain a literal for atom $p_k$, $\frac{\partial (1-\mathbf{L}_v[i,k])}{\partial \mathbf{v}[k]} = 0$ and $\frac{\partial \mathbf{L}_v[i,k]}{\partial \mathbf{v}[k]} = 0$;
	\item [{\bf (c)}] when clause $i$ contains literal $p_k$, $\frac{\partial (1-\mathbf{L}_v[i,k])}{\partial \mathbf{v}[k]} = -1$ and $\frac{\partial \mathbf{L}_v[i,k]}{\partial \mathbf{v}[k]} = 1$;
	\item [{\bf (d)}] when clause $i$ contains literal $\neg p_k$, $\frac{\partial (1-\mathbf{L}_v[i,k])}{\partial \mathbf{v}[k]} = 1$ and $\frac{\partial \mathbf{L}_v[i,k]}{\partial \mathbf{v}[k]} = -1$.
\end{itemize}
We will refer to the above 4 bullets with their identifiers.

Since $i\in U$, we know clause $i$ has all but one literal of the form $\neg p_j$ such that $p_j \in F$. 
Since $\mathbf{f}[k]=0$, we know $p_k\not \in F$. Then, when clause $i$ contains literal $p_k$ or $\neg p_k$, all other literals in clause $i$ must be of the form $\neg p_j$ where $p_j \in F$. 
For every literal $\neg p_j$ in clause $i$ where $j\neq k$, we know $p_j \in F$, thus $\mathbf{f}[j]=1$; since $\mathbf{v} = \mathbf{f} + \mathbbm{1}_{\{0 \}}(\mathbf{f}) \odot b_p(\xx)$, then $\mathbf{v}[j]=1$; consequently, the literal $\neg p_j$ evaluates to $\false$ under $v$. Recall that $\mathbf{L}_v[i,j]\in\{0,1\}$, and $\mathbf{L}_v[i,j]=1$ iff clause $i$ contains a literal ($p_j$ or $\neg p_j$) for atom $p_j$ and this literal evaluates to $\true$ under $v$, then we know
\begin{itemize}
	\item when $i\in U$, $\mathbf{f}[k]=0$, and clause $i$ contains literal $p_k$ or $\neg p_k$, $\mathbf{L}_v[i,j]=0$ for $j\in \{1,\dots,n\}$ such that $j\neq k$.
\end{itemize}
Then we have
{\small
	\begin{align*}
		&G_{i,k}\\
		=& \left(\prod\limits_{j\in\{1,\dots,n\}}(1-\mathbf{L}_v[i,j])\right)
		\times \sum\limits_{j\in\{1,\dots,n\}}
		\frac{\frac{\partial (1-\mathbf{L}_v[i,j])}{\partial \mathbf{v}[k]}}{1-\mathbf{L}_v[i,j]}\\
		=& \left(\prod\limits_{j\in\{1,\dots,n\}}(1-\mathbf{L}_v[i,j])\right)
		\times \frac{\frac{\partial (1-\mathbf{L}_v[i,k])}{\partial \mathbf{v}[k]}}{1-\mathbf{L}_v[i,k]} ~\text{(due to {\bf (a)})}\\
		=& \frac{\partial (1-\mathbf{L}_v[i,k])}{\partial \mathbf{v}[k]} \times \prod\limits_{\substack{j\in\{1,\dots,n\}\\j\neq k}}(1-\mathbf{L}_v[i,j])\\
		=&
		\begin{cases}
			0 & \text{if clause $i$ doesn't contain a literal} \\
			& \text{for atom $p_k$}~\text{(due to {\bf (b)})} \\
			-1& \text{if clause $i$ contains a literal $p_k$} ~\text{(due to {\bf (c)})} \\
			1& \text{if clause $i$ contains a literal $\neg p_k$} ~\text{(due to {\bf (d)})}
		\end{cases}
	\end{align*}
}%
Since $i\in U$ and $\mathbf{f}[k]=0$, when clause $i$ contains a literal $l_k$ for atom $p_k$, we know $F\not \models l_j$ for every literal $l_j$ in clause $i$ such that $j\neq k$. Since $C\cup F$ is satisfiable, we know $C\cup F \models l_k$ and there cannot be two clauses in $C_{deduce}$ containing different literals $p_k$ and $\neg p_k$. Thus, when $\mathbf{f}[k]=0$,
{\small
	\begin{align*}
		&\frac{\partial L_{deduce}}{\partial \mathbf{v}[k]}\\
		=& \sum\limits_{i\in U}G_{i,k}\\
		=&
		\begin{cases}
			-c& \text{if $c>0$ clauses in $C_{deduce}$ contain literal $p_k$,} \\
			c& \text{if $c>0$ clauses in $C_{deduce}$ contain literal $\neg p_k$,}\\
			0 & \text{otherwise.}
		\end{cases}
	\end{align*}
}%
Note that the first 2 cases above are disjoint since there cannot be two clauses in $C_{deduce}$ containing different literals $p_k$ and $\neg p_k$.

Finally, if $p_k\not \in F$,
\begin{align*}
	&\frac{\partial L_{deduce}}{\partial \xx[k]}\\
	\stackrel{iSTE}{\approx}& \frac{\partial L_{deduce}}{\partial \mathbf{v}[k]}\\
	=~~&
	\begin{cases}
		-c & \text{if $c>0$ clauses in $C_{deduce}$} \\
		&\text{contain literal $p_k$;}\\
		c & \text{if $c>0$ clauses in $C_{deduce}$}\\
		&\text{contain literal $\neg p_k$;} \\
		0 & \text{otherwise;}
	\end{cases}
\end{align*}

\medskip
[$L_{unsat}$]~~
According to the definition,
\begin{align*}
	L_{unsat} =& avg(\mathbbm{1}_{\{1\}}(\mathbf{unsat}) \odot \mathbf{unsat} )\\
	=& \frac{1}{m}\sum\limits_{i\in \{1,\dots,m\}}\Big( \mathbbm{1}_{\{1\}}(\mathbf{unsat}[i]) \times \\
	& \hspace{2cm} \prod\limits_{j\in\{1,\dots,n\}}(1-\mathbf{L}_v[i,j]) \Big)
\end{align*}
Recall that we proved that $\mathbbm{1}_{\{1\}}(\mathbf{unsat})[i]\in \{0,1\}$ is the output of an indicator function whose value is $1$ iff clause $i$ evaluates to $\false$ under $v$. Let $U\subseteq \{1,\dots,m\}$ denote the set of indices of clauses in $C$ that are evaluated as $\false$ under $v$.
\begin{align*}
	L_{unsat} =& \frac{1}{m}\sum\limits_{i\in U}\Big( \prod\limits_{j\in\{1,\dots,n\}}(1-\mathbf{L}_v[i,j]) \Big)
\end{align*}
Then the gradient of $L_{unsat}$ w.r.t. $\mathbf{v}[k]$ is
\begin{align*}
	\frac{\partial L_{unsat}}{\partial \mathbf{v}[k]} =& \frac{1}{m}\sum\limits_{i\in U}\Big(\frac{\partial \prod\limits_{j\in\{1,\dots,n\}}(1-\mathbf{L}_v[i,j])}{\partial \mathbf{v}[k]} \Big).
\end{align*}
Recall that $\mathbf{L}_v[i,j]\in\{0,1\}$, and $\mathbf{L}_v[i,j]=1$ iff clause $i$ contains a literal ($p_j$ or $\neg p_j$) for atom $p_j$ and this literal evaluates to $\true$ under $v$.
When $i\in U$, clause $i$ evaluates to $\false$ under $v$. Thus when $i\in U$, all literals in clause $i$ must be evaluated as $\false$ under $v$, and consequently, $\mathbf{L}_v[i,j]=0$ for all $j\in \{1,\dots,m\}$. Then
\begin{align*}
	\frac{\partial L_{unsat}}{\partial \mathbf{v}[k]} =& \frac{1}{m}\sum\limits_{i\in U}\Big(\frac{\partial \prod\limits_{j\in\{1,\dots,n\}}(1-\mathbf{L}_v[i,j])}{\partial \mathbf{v}[k]} \Big)\\
	=& \frac{1}{m}\sum\limits_{i\in U}\Big(\frac{\partial (1-\mathbf{L}_v[i,k])}{\partial \mathbf{v}[k]} \Big) ~\text{(due to {\bf (a)})}\\
	=& \frac{c_2-c_1}{m} ~\text{(due to {\bf (b), (c), (d)})}
\end{align*}
where $c_1$ (and $c_2$, resp.) is the number of clauses in $U$ that contain $p_k$ (and $\neg p_k$, resp.). 
Finally, if $p_k\not \in F$,
\begin{align*}
	\frac{\partial L_{unsat}}{\partial \xx[k]} 
	\stackrel{iSTE}{\approx} \frac{\partial L_{unsat}}{\partial \mathbf{v}[k]} 
	= \frac{c_2 - c_1}{m}
\end{align*}
where $c_1$ (and $c_2$, resp.) is the number of clauses in $C$ that are not satisfied by $v$ and contain $p_k$ (and $\neg p_k$, resp.).

\medskip
[$L_{sat}$]~~
Recall that we proved that $\mathbbm{1}_{\{0\}}(\mathbf{unsat})[i]\in \{0,1\}$ is the output of an indicator function whose value is $1$ iff clause $i$ evaluates to $\true$ under $v$. Let $S\subseteq \{1,\dots,m\}$ denote the set of indices of clauses in $C$ that are evaluated as $\true$ under $v$. Then
{\small
\begin{align*}
	&L_{sat} \\
	=& avg(\mathbbm{1}_{\{0\}}(\mathbf{unsat}) \odot \mathbf{keep} )\\
	=& \frac{1}{m}\sum\limits_{i\in \{1,\dots,m\}}\Big( \mathbbm{1}_{\{0\}}(\mathbf{unsat}[i]) \times \mathbf{keep}[i] \Big)\\
	=& \frac{1}{m}\sum\limits_{i\in S}\mathbf{keep}[i] \\
	=& \frac{1}{m}
	\sum\limits_{i\in S}
	\sum\limits_{j\in \{1,\dots,n\}} 
	\Big(
	\mathbbm{1}_{\{1\}}(\mathbf{L}_v[i,j]) \times (1-\mathbf{L}_v[i,j])\\
	& \hspace{2.2cm} + \mathbbm{1}_{\{0\}}(\mathbf{L}_v[i,j]) \times \mathbf{L}_v[i,j]
	\Big)
\end{align*}
}
Then the gradient of $L_{sat}$ w.r.t. $\mathbf{v}[k]$ is
{\small 
	\begin{align*}
		&\frac{\partial L_{sat}}{\partial \mathbf{v}[k]} \\
		=& \frac{1}{m}
		\sum\limits_{i\in S}
		\sum\limits_{j\in \{1,\dots,n\}} 
		\Big(\\
		& \hspace{1cm} \mathbbm{1}_{\{1\}}(\mathbf{L}_v[i,j]) \times 
		\frac{\partial (1-\mathbf{L}_v[i,j])}{\partial \mathbf{v}[k]}
		\\
		& \hspace{1cm} + \mathbbm{1}_{\{0\}}(\mathbf{L}_v[i,j]) \times 
		\frac{\partial \mathbf{L}_v[i,j]}{\partial \mathbf{v}[k]}
		\Big)
	\end{align*}
	\begin{align*}
		=& \frac{1}{m}
		\sum\limits_{i\in S}
		\Big(
		\mathbbm{1}_{\{1\}}(\mathbf{L}_v[i,k]) \times 
		\frac{\partial (1-\mathbf{L}_v[i,k])}{\partial \mathbf{v}[k]}
		\\
		& \hspace{1cm} + \mathbbm{1}_{\{0\}}(\mathbf{L}_v[i,k]) \times 
		\frac{\partial \mathbf{L}_v[i,k]}{\partial \mathbf{v}[k]}
		\Big)~\text{(due to {\bf (a)})}
	\end{align*}
	\begin{align*}
		=& \frac{1}{m}
		\sum\limits_{\substack{i\in S\\\text{clause $i$ contains}\\\text{literal $p_k$}}}
		\Big(
		\mathbbm{1}_{\{1\}}(\mathbf{L}_v[i,k]) \times 
		\frac{\partial (1-\mathbf{L}_v[i,k])}{\partial \mathbf{v}[k]}
		\\
		& \hspace{1cm} + \mathbbm{1}_{\{0\}}(\mathbf{L}_v[i,k]) \times 
		\frac{\partial \mathbf{L}_v[i,k]}{\partial \mathbf{v}[k]}
		\Big) + \\
		& \frac{1}{m}
		\sum\limits_{\substack{i\in S\\\text{clause $i$ contains}\\\text{literal $\neg p_k$}}}
		\Big(
		\mathbbm{1}_{\{1\}}(\mathbf{L}_v[i,k]) \times 
		\frac{\partial (1-\mathbf{L}_v[i,k])}{\partial \mathbf{v}[k]}
		\\
		& \hspace{1cm} + \mathbbm{1}_{\{0\}}(\mathbf{L}_v[i,k]) \times 
		\frac{\partial \mathbf{L}_v[i,k]}{\partial \mathbf{v}[k]}
		\Big)~\text{(due to {\bf (b)})}
	\end{align*}
	\begin{align*}
		=& \frac{1}{m}
		\sum\limits_{\substack{i\in S\\\text{clause $i$ contains}\\\text{literal $p_k$}}}
		\Big(
		-\mathbbm{1}_{\{1\}}(\mathbf{L}_v[i,k])
		+ \mathbbm{1}_{\{0\}}(\mathbf{L}_v[i,k])
		\Big)\\
		& +\frac{1}{m}
		\sum\limits_{\substack{i\in S\\\text{clause $i$ contains}\\\text{literal $\neg p_k$}}}
		\Big(
		\mathbbm{1}_{\{1\}}(\mathbf{L}_v[i,k])
		- \mathbbm{1}_{\{0\}}(\mathbf{L}_v[i,k])
		\Big)\\
		& ~\text{(due to {\bf (c)} and {\bf (d)})}
	\end{align*}
}%
Recall that $\mathbf{L}_v[i,j]\in\{0,1\}$, and $\mathbf{L}_v[i,j]=1$ iff clause $i$ contains a literal ($p_j$ or $\neg p_j$) for atom $p_j$ and this literal evaluates to $\true$ under $v$.
It's easy to check that
\begin{itemize}
	\item when clause $i$ contains literal $p_k$, the value of $-\mathbbm{1}_{\{1\}}(\mathbf{L}_v[i,k])
	+ \mathbbm{1}_{\{0\}}(\mathbf{L}_v[i,k])$ is $-1$ if $v\models p_k$ and is $1$ if $v\not\models p_k$;
	\item when clause $i$ contains literal $\neg p_k$, the value of $\mathbbm{1}_{\{1\}}(\mathbf{L}_v[i,k])
	- \mathbbm{1}_{\{0\}}(\mathbf{L}_v[i,k])$ is $-1$ if $v\models p_k$ and is $1$ if $v\not\models p_k$.
\end{itemize}
Thus
\begin{align*}
	\frac{\partial L_{sat}}{\partial \mathbf{v}[k]} =& 
	\begin{cases}
		-\frac{c}{m} & \text{if $v\models p_k$,}\\
		\frac{c}{m} & \text{if $v\not \models p_k$.}
	\end{cases}
\end{align*}
where $c$ is the number of clauses in $S$ that contain a literal for atom $p_k$.
Finally, if $p_k\not \in F$,
\begin{align*}
	\frac{\partial L_{sat}}{\partial \xx[k]} 
	\stackrel{iSTE}{\approx}  \frac{\partial L_{sat}}{\partial \mathbf{v}[k]}
	=
	\begin{cases}
		-\frac{c}{m} & \text{if $v\models p_k$,} \\
		\frac{c}{m} & \text{if $v\not \models p_k$;} 
	\end{cases}
\end{align*}
where $c$ is the number of clauses in $C$ that are satisfied by $v$ and contain $p_k$ or $\neg p_k$.
\end{enumerate}
\end{proof}

\bigskip
\noindent{\bf Proposition~\ref{prop:gradient}~~}
Proposition~\ref{prop:gradient:b_p} still holds for
{${\bf x}\in \mathbb{R}^n$}
and $\mathbf{v} = \mathbf{f} + \mathbbm{1}_{\{0 \}}(\mathbf{f}) \odot b(\xx)$.

[Complete Statement] 
Given a CNF theory $C$ of $m$ clauses and $n$ atoms and a set $F$ of atoms such that $C\cup F$ is satisfiable, let $\mathbf{C}, \mathbf{f}$ denote their matrix/vector representations, respectively. 
Given a neural network output $\mathbf{x}\in \mathbb{R}^{n}$ in logits (i.e., real numbers instead of probabilities),
we construct $\mathbf{v} = \mathbf{f} + \mathbbm{1}_{\{0 \}}(\mathbf{f}) \odot b(\xx)$ 
and a truth assignment $v$ such that $v(p_j)=\true$ if $\mathbf{v}[j]$ is $1$, and $v(p_j)=\false$ if $\mathbf{v}[j]$ is $0$. 
Let $C_{deduce} \subseteq C$ denote the set of Horn clauses $H$ in $C$ such that all but one literal in $H$ are of the form $\neg p$ and $p\in F$.
Then, for any $j\in \{1,\dots,n\}$,
\begin{enumerate}
\item if $p_j\in F$, all of $\frac{\partial L_{deduce}}{\partial \xx[j]}$, $\frac{\partial L_{unsat}}{\partial \xx[j]}$, and $\frac{\partial L_{sat}}{\partial \xx[j]}$ are zeros;
\item if $p_j\not \in F$,
\begin{align*}
\frac{\partial L_{deduce}}{\partial \xx[j]}& \stackrel{iSTE}{\approx}
\begin{cases}
	-c & \text{if $c>0$ clauses in $C_{deduce}$} \\
	&\text{contain literal $p_j$;}\\
	c & \text{if $c>0$ clauses in $C_{deduce}$}\\
	&\text{contain literal $\neg p_j$;} \\
	0 & \text{otherwise;}
\end{cases} \\
\frac{\partial L_{unsat}}{\partial \xx[j]} &
\stackrel{iSTE}{\approx}
\frac{c_2 - c_1}{m} \\
\frac{\partial L_{sat}}{\partial \xx[j]} &
\stackrel{iSTE}{\approx}
\begin{cases}
	-\frac{c_3}{m} & \text{if $v\models p_j$,} \\
	\frac{c_3}{m} & \text{if $v\not \models p_j$.} 
\end{cases}
\end{align*}
where $\stackrel{iSTE}{\approx}$ stands for the equivalence of gradients under iSTE; $c_1$ (and $c_2$, resp.) is the number of clauses in $C$ that are not satisfied by $v$ and contain $p_j$ (and $\neg p_j$, resp.); $c_3$ is the number of clauses in $C$ that are satisfied by $v$ and contain $p_j$ or $\neg p_j$.
\end{enumerate}

\begin{proof}
Recall the definition of $L_{cnf}$
{\small
\begin{align*}
\mathbf{L}_f =\ & \mathbf{C} \odot \mathbf{f}  \\
\mathbf{L}_v =\ & \mathbbm{1}_{\{1\}}(\mathbf{C}) \odot \mathbf{v} + \mathbbm{1}_{\{-1\}}(\mathbf{C}) \odot (1-\mathbf{v}) \\
\mathbf{deduce} =\ & \mathbbm{1}_{\{1\}}\Big(sum(\mathbf{C} \odot \mathbf{C}) - sum(\mathbbm{1}_{\{-1\}}(\mathbf{L}_f)) \Big)  \\
\mathbf{unsat} =\ & prod(1-\mathbf{L}_v)  \\
\mathbf{keep} =\ & sum(\mathbbm{1}_{\{1\}}(\mathbf{L}_v) \odot (1-\mathbf{L}_v) + \mathbbm{1}_{\{0\}}(\mathbf{L}_v) \odot \mathbf{L}_v) \\
L_{deduce} =\ & sum(\mathbf{deduce} \odot \mathbf{unsat}) \\
L_{unsat} =\ & avg( \mathbbm{1}_{\{1\}}(\mathbf{unsat}) \odot \mathbf{unsat} )  \\
L_{sat} =\ & avg(\mathbbm{1}_{\{0\}}(\mathbf{unsat}) \odot \mathbf{keep} )
\end{align*}
\begin{align*}
L_{cnf}(\mathbf{C}, \mathbf{v}, \mathbf{f}) =\ & L_{deduce} + L_{unsat} + L_{sat}
\end{align*}
}
We will prove each bullet in Proposition~\ref{prop:gradient} as follows. This proof is almost the same as the proof for Proposition~\ref{prop:gradient:b_p} since the choice of $b(x)$ v.s. $b_p(x)$ doesn't affect the gradient computation from $L_{cnf}$ to $\xx$ under iSTE.
\begin{enumerate}
\item 
Take any $k\in\{1,\dots,n\}$, let's focus on $\xx[k]$ and compute the gradient of $L\in \{L_{deduce},L_{unsat},L_{sat} \}$ to it with iSTE. According to the chain rule and since $\frac{\partial \mathbf{v}[i]}{\partial b(\xx)[j]}=0$ for $i\neq j$, we have
\begin{align*}
	\frac{\partial L}{\partial \xx[k]}
	=& \frac{\partial L}{\partial \mathbf{v}[k]} 
	\times \frac{\partial \mathbf{v}[k]}{\partial b(\xx[k])}
	\times \frac{\partial b(\xx[k])}{\partial \xx[k]}.
\end{align*}
Under iSTE, the last term $\frac{\partial b(\xx[k])}{\partial \xx[k]}$ is replaced with $\frac{\partial s(\xx[k])}{\partial \xx[k]} = \frac{\partial \xx[k]}{\partial \xx[k]}=1$. Thus
\begin{align*}
	\frac{\partial L}{\partial \xx[k]}
	=& \frac{\partial L}{\partial \mathbf{v}[k]} 
	\times \frac{\partial \mathbf{v}[k]}{\partial b(\xx[k])}
	\times \frac{\partial b(\xx[k])}{\partial \xx[k]}\\
	\stackrel{iSTE}{\approx} & \frac{\partial L}{\partial \mathbf{v}[k]} 
	\times \frac{\partial \mathbf{v}[k]}{\partial b(\xx[k])} ~~\text{(under iSTE)}\\
	=& \frac{\partial L}{\partial \mathbf{v}[k]} 
	\times \frac{\partial (\mathbf{f}[k] + \mathbbm{1}_{\{0 \}}(\mathbf{f}[k]) \times b(\xx[k]))}{\partial b(\xx[k])}\\
	=&
	\begin{cases}
		\frac{\partial L}{\partial \mathbf{v}[k]} & \text{if $\mathbf{f}[k]=0$,}\\
		0 & \text{if $\mathbf{f}[k]=1$.}
	\end{cases}
\end{align*}
Since $\mathbf{f}[k]=1$ iff $p_k\in F$, if $p_k\in F$, then all of $\frac{\partial L_{deduce}}{\partial \xx[k]}$, $\frac{\partial L_{unsat}}{\partial \xx[k]}$, and $\frac{\partial L_{sat}}{\partial \xx[k]}$ are zeros.
\item 
We know $p_k\not\in F$ iff $\mathbf{f}[k]=0$. As proved in the first bullet, for $L\in \{L_{deduce},L_{unsat},L_{sat} \}$, if $p_k\not\in F$, then $\frac{\partial L}{\partial \xx[k]} = \frac{\partial L}{\partial \mathbf{v}[k]}$. We further analyze the value of $\frac{\partial L}{\partial \mathbf{v}[k]}$ for each $L$ under the condition that $\mathbf{f}[k]=0$.

\medskip
[$L_{deduce}$]~~According to the definition,
{\small
	\begin{align*}
		&L_{deduce} \\
		=& \sum\limits_{i\in \{1,\dots,m\}}\Big( \mathbf{deduce}[i] \times \mathbf{unsat}[i] \Big)\\
		=& \sum\limits_{i\in \{1,\dots,m\}}\Big( \mathbf{deduce}[i] \times \prod\limits_{j\in\{1,\dots,n\}}(1-\mathbf{L}_v[i,j]) \Big)
	\end{align*}
	\begin{align*}
		&\frac{\partial L_{deduce}}{\partial \mathbf{v}[k]} \\
		=& \sum\limits_{i\in \{1,\dots,m\}}\frac{\partial \Big( \mathbf{deduce}[i] \times \prod\limits_{j\in\{1,\dots,n\}}(1-\mathbf{L}_v[i,j])\Big)}{\partial \mathbf{v}[k]}\\
		=& \sum\limits_{i\in \{1,\dots,m\}}\Big(
		\frac{\partial \mathbf{deduce}[i]}{\partial \mathbf{v}[k]} \times \prod\limits_{j\in\{1,\dots,n\}}(1-\mathbf{L}_v[i,j]) +\\
		& \hspace{1.8cm} \mathbf{deduce}[i] \times
		\frac{\partial \prod\limits_{j\in\{1,\dots,n\}}(1-\mathbf{L}_v[i,j])}{\partial \mathbf{v}[k]} \Big)
	\end{align*}
}%
Since $\mathbf{deduce}$ is the result of an indicator function, $\frac{\partial \mathbf{deduce}[i]}{\partial \mathbf{v}[k]}=0$. Then,
{\small
	\begin{align*}
		&\frac{\partial L_{deduce}}{\partial \mathbf{v}[k]} \\
		=& \sum\limits_{i\in \{1,\dots,m\}}\Big(\mathbf{deduce}[i] \times
		\frac{\partial \prod\limits_{j\in\{1,\dots,n\}}(1-\mathbf{L}_v[i,j])}{\partial \mathbf{v}[k]} \Big).
	\end{align*}
}%
Let $U\subseteq \{1,\dots,m\}$ denote the set of indices of all clauses in $C_{deduce}$. Since $\mathbf{deduce}[i]=1$ iff $i\in U$,
\begin{align*}
	\frac{\partial L_{deduce}}{\partial \mathbf{v}[k]} =& \sum\limits_{i\in U}\Big(\frac{\partial \prod\limits_{j\in\{1,\dots,n\}}(1-\mathbf{L}_v[i,j])}{\partial \mathbf{v}[k]} \Big).
\end{align*}
Let $G_{i,k}$ denote $\frac{\partial \prod\limits_{j\in\{1,\dots,n\}}(1-\mathbf{L}_v[i,j])}{\partial \mathbf{v}[k]}$. Then
\begin{align*}
	\frac{\partial L_{deduce}}{\partial \mathbf{v}[k]} =& \sum\limits_{i\in U}G_{i,k}.
\end{align*}
Let's analyze the value of $G_{i,k}$ where $i\in U$ and $k\in \{1,\dots,n\}$ such that $\mathbf{f}[k]=0$.
According to the product rule below, 
\begin{align*}
	\frac{d}{d x}\left[\prod_{i=1}^{k} f_{i}(x)\right]=\left(\prod_{i=1}^{k} f_{i}(x)\right)\left(\sum_{i=1}^{k} \frac{f_{i}^{\prime}(x)}{f_{i}(x)}\right)
\end{align*}
we have
{\small
	\begin{align*}
		&G_{i,k}\\
		=& \frac{\partial \prod\limits_{j\in\{1,\dots,n\}}(1-\mathbf{L}_v[i,j])}{\partial \mathbf{v}[k]} \\
		=& \left(\prod\limits_{j\in\{1,\dots,n\}}(1-\mathbf{L}_v[i,j])\right)
		\times \sum\limits_{j\in\{1,\dots,n\}}
		\frac{\frac{\partial (1-\mathbf{L}_v[i,j])}{\partial \mathbf{v}[k]}}{1-\mathbf{L}_v[i,j]}
	\end{align*}
}%
Since $\mathbf{L}_v[i,j]=\mathbbm{1}_{\{1\}}(\mathbf{C})[i,j] \times \mathbf{v}[j] + \mathbbm{1}_{\{-1\}}(\mathbf{C})[i,j] \times (1-\mathbf{v}[j])$, we know
\begin{itemize}
	\item [{\bf (a)}] for $j\in\{1,\dots,n\}$ such that $j\neq k$, $\frac{\partial (1-\mathbf{L}_v[i,j])}{\partial \mathbf{v}[k]} = 0$ and $\frac{\partial \mathbf{L}_v[i,j]}{\partial \mathbf{v}[k]} = 0$;
	\item [{\bf (b)}] when clause $i$ doesn't contain a literal for atom $p_k$, $\frac{\partial (1-\mathbf{L}_v[i,k])}{\partial \mathbf{v}[k]} = 0$ and $\frac{\partial \mathbf{L}_v[i,k]}{\partial \mathbf{v}[k]} = 0$;
	\item [{\bf (c)}] when clause $i$ contains literal $p_k$, $\frac{\partial (1-\mathbf{L}_v[i,k])}{\partial \mathbf{v}[k]} = -1$ and $\frac{\partial \mathbf{L}_v[i,k]}{\partial \mathbf{v}[k]} = 1$;
	\item [{\bf (d)}] when clause $i$ contains literal $\neg p_k$, $\frac{\partial (1-\mathbf{L}_v[i,k])}{\partial \mathbf{v}[k]} = 1$ and $\frac{\partial \mathbf{L}_v[i,k]}{\partial \mathbf{v}[k]} = -1$.
\end{itemize}
We will refer to the above 4 bullets with their identifiers.

Since $i\in U$, we know clause $i$ has all but one literal of the form $\neg p_j$ such that $p_j \in F$. 
Since $\mathbf{f}[k]=0$, we know $p_k\not \in F$. Then, when clause $i$ contains literal $p_k$ or $\neg p_k$, all other literals in clause $i$ must be of the form $\neg p_j$ where $p_j \in F$. 
For every literal $\neg p_j$ in clause $i$ where $j\neq k$, we know $p_j \in F$, thus $\mathbf{f}[j]=1$; since $\mathbf{v} = \mathbf{f} + \mathbbm{1}_{\{0 \}}(\mathbf{f}) \odot b(\xx)$, then $\mathbf{v}[j]=1$; consequently, the literal $\neg p_j$ evaluates to $\false$ under $v$. Recall that $\mathbf{L}_v[i,j]\in\{0,1\}$, and $\mathbf{L}_v[i,j]=1$ iff clause $i$ contains a literal ($p_j$ or $\neg p_j$) for atom $p_j$ and this literal evaluates to $\true$ under $v$, then we know
\begin{itemize}
	\item when $i\in U$, $\mathbf{f}[k]=0$, and clause $i$ contains literal $p_k$ or $\neg p_k$, $\mathbf{L}_v[i,j]=0$ for $j\in \{1,\dots,n\}$ such that $j\neq k$.
\end{itemize}
Then we have
{\small 
	\begin{align*}
		&G_{i,k}\\
		=& \left(\prod\limits_{j\in\{1,\dots,n\}}(1-\mathbf{L}_v[i,j])\right)
		\times \sum\limits_{j\in\{1,\dots,n\}}
		\frac{\frac{\partial (1-\mathbf{L}_v[i,j])}{\partial \mathbf{v}[k]}}{1-\mathbf{L}_v[i,j]}\\
		=& \left(\prod\limits_{j\in\{1,\dots,n\}}(1-\mathbf{L}_v[i,j])\right)
		\times \frac{\frac{\partial (1-\mathbf{L}_v[i,k])}{\partial \mathbf{v}[k]}}{1-\mathbf{L}_v[i,k]} ~\text{(due to {\bf (a)})}\\
		=& \frac{\partial (1-\mathbf{L}_v[i,k])}{\partial \mathbf{v}[k]} \times \prod\limits_{\substack{j\in\{1,\dots,n\}\\j\neq k}}(1-\mathbf{L}_v[i,j])\\
		=&
		\begin{cases}
			0 & \text{if clause $i$ doesn't contain a literal} \\
			& \text{for atom $p_k$}~\text{(due to {\bf (b)})} \\
			-1& \text{if clause $i$ contains a literal $p_k$} ~\text{(due to {\bf (c)})} \\
			1& \text{if clause $i$ contains a literal $\neg p_k$} ~\text{(due to {\bf (d)})}
		\end{cases}
	\end{align*}
}%
Since $i\in U$ and $\mathbf{f}[k]=0$, when clause $i$ contains a literal $l_k$ for atom $p_k$, we know $F\not \models l_j$ for every literal $l_j$ in clause $i$ such that $j\neq k$. Since $C\cup F$ is satisfiable, we know $C\cup F \models l_k$ and there cannot be two clauses in $C_{deduce}$ containing different literals $p_k$ and $\neg p_k$. Thus, when $\mathbf{f}[k]=0$,
{\small
\begin{align*}
	&\frac{\partial L_{deduce}}{\partial \mathbf{v}[k]} \\
	=& \sum\limits_{i\in U}G_{i,k}\\
	=&
	\begin{cases}
		-c& \text{if $c>0$ clauses in $C_{deduce}$ contain literal $p_k$,} \\
		c& \text{if $c>0$ clauses in $C_{deduce}$ contain literal $\neg p_k$,}\\
		0 & \text{otherwise.}
	\end{cases}
\end{align*}
}
Note that the first 2 cases above are disjoint since there cannot be two clauses in $C_{deduce}$ containing different literals $p_k$ and $\neg p_k$.

Finally, if $p_k\not \in F$,
{\small 
\begin{align*}
	&\frac{\partial L_{deduce}}{\partial \xx[k]}\\
	\stackrel{iSTE}{\approx}& \frac{\partial L_{deduce}}{\partial \mathbf{v}[k]}\\
	=~~&
	\begin{cases}
		-c & \text{if $c>0$ clauses in $C_{deduce}$} \\
		&\text{contain literal $p_k$;}\\
		c & \text{if $c>0$ clauses in $C_{deduce}$}\\
		&\text{contain literal $\neg p_k$;} \\
		0 & \text{otherwise;}
	\end{cases}
\end{align*}
}

\medskip
[$L_{unsat}$]~~
According to the definition,
\begin{align*}
	L_{unsat} =& avg(\mathbbm{1}_{\{1\}}(\mathbf{unsat}) \odot \mathbf{unsat} )\\
	=& \frac{1}{m}\sum\limits_{i\in \{1,\dots,m\}}\Big( \mathbbm{1}_{\{1\}}(\mathbf{unsat}[i]) \times \\
	& \hspace{2cm} \prod\limits_{j\in\{1,\dots,n\}}(1-\mathbf{L}_v[i,j]) \Big)
\end{align*}
Recall that we proved that $\mathbbm{1}_{\{1\}}(\mathbf{unsat})[i]\in \{0,1\}$ is the output of an indicator function whose value is $1$ iff clause $i$ evaluates to $\false$ under $v$. Let $U\subseteq \{1,\dots,m\}$ denote the set of indices of clauses in $C$ that are evaluated as $\false$ under $v$.
\begin{align*}
	L_{unsat} =& \frac{1}{m}\sum\limits_{i\in U}\Big( \prod\limits_{j\in\{1,\dots,n\}}(1-\mathbf{L}_v[i,j]) \Big)
\end{align*}
Then the gradient of $L_{unsat}$ w.r.t. $\mathbf{v}[k]$ is
\begin{align*}
	\frac{\partial L_{unsat}}{\partial \mathbf{v}[k]} =& \frac{1}{m}\sum\limits_{i\in U}\Big(\frac{\partial \prod\limits_{j\in\{1,\dots,n\}}(1-\mathbf{L}_v[i,j])}{\partial \mathbf{v}[k]} \Big).
\end{align*}
Recall that $\mathbf{L}_v[i,j]\in\{0,1\}$, and $\mathbf{L}_v[i,j]=1$ iff clause $i$ contains a literal ($p_j$ or $\neg p_j$) for atom $p_j$ and this literal evaluates to $\true$ under $v$.
When $i\in U$, clause $i$ evaluates to $\false$ under $v$. Thus when $i\in U$, all literals in clause $i$ must be evaluated as $\false$ under $v$, and consequently, $\mathbf{L}_v[i,j]=0$ for all $j\in \{1,\dots,m\}$. Then
\begin{align*}
	\frac{\partial L_{unsat}}{\partial \mathbf{v}[k]} =& \frac{1}{m}\sum\limits_{i\in U}\Big(\frac{\partial \prod\limits_{j\in\{1,\dots,n\}}(1-\mathbf{L}_v[i,j])}{\partial \mathbf{v}[k]} \Big)\\
	=& \frac{1}{m}\sum\limits_{i\in U}\Big(\frac{\partial (1-\mathbf{L}_v[i,k])}{\partial \mathbf{v}[k]} \Big) ~\text{(due to {\bf (a)})}\\
	=& \frac{c_2-c_1}{m}
\end{align*}
where $c_1$ (and $c_2$, resp.) is the number of clauses in $U$ that contain $p_k$ (and $\neg p_k$, resp.). 
Finally, if $p_k\not \in F$,
\begin{align*}
	\frac{\partial L_{unsat}}{\partial \xx[k]} 
	\stackrel{iSTE}{\approx} \frac{\partial L_{unsat}}{\partial \mathbf{v}[k]} 
	= \frac{c_2 - c_1}{m}
\end{align*}
where $c_1$ (and $c_2$, resp.) is the number of clauses in $C$ that are not satisfied by $v$ and contain $p_k$ (and $\neg p_k$, resp.).

\medskip
[$L_{sat}$]~~
Recall that we proved that $\mathbbm{1}_{\{0\}}(\mathbf{unsat})[i]\in \{0,1\}$ is the output of an indicator function whose value is $1$ iff clause $i$ evaluates to $\true$ under $v$. Let $S\subseteq \{1,\dots,m\}$ denote the set of indices of clauses in $C$ that are evaluated as $\true$ under $v$. Then
{\small
\begin{align*}
	L_{sat} =& avg(\mathbbm{1}_{\{0\}}(\mathbf{unsat}) \odot \mathbf{keep} )\\
	=& \frac{1}{m}\sum\limits_{i\in \{1,\dots,m\}}\Big( \mathbbm{1}_{\{0\}}(\mathbf{unsat}[i]) \times \mathbf{keep}[i] \Big)\\
	=& \frac{1}{m}\sum\limits_{i\in S}\mathbf{keep}[i] \\
	=& \frac{1}{m}
	\sum\limits_{i\in S}
	\sum\limits_{j\in \{1,\dots,n\}} 
	\Big(
	\mathbbm{1}_{\{1\}}(\mathbf{L}_v[i,j]) \times (1-\mathbf{L}_v[i,j])\\
	& \hspace{2.2cm} + \mathbbm{1}_{\{0\}}(\mathbf{L}_v[i,j]) \times \mathbf{L}_v[i,j]
	\Big)
\end{align*}
}
Then the gradient of $L_{sat}$ w.r.t. $\mathbf{v}[k]$ is
{\small 
	\begin{align*}
		&\frac{\partial L_{sat}}{\partial \mathbf{v}[k]} \\
		=& \frac{1}{m}
		\sum\limits_{i\in S}
		\sum\limits_{j\in \{1,\dots,n\}} 
		\Big(\\
		& \hspace{1cm} \mathbbm{1}_{\{1\}}(\mathbf{L}_v[i,j]) \times 
		\frac{\partial (1-\mathbf{L}_v[i,j])}{\partial \mathbf{v}[k]}
		\\
		& \hspace{1cm} + \mathbbm{1}_{\{0\}}(\mathbf{L}_v[i,j]) \times 
		\frac{\partial \mathbf{L}_v[i,j]}{\partial \mathbf{v}[k]}
		\Big)
	\end{align*}
	\begin{align*}
		=& \frac{1}{m}
		\sum\limits_{i\in S}
		\Big(
		\mathbbm{1}_{\{1\}}(\mathbf{L}_v[i,k]) \times 
		\frac{\partial (1-\mathbf{L}_v[i,k])}{\partial \mathbf{v}[k]}
		\\
		& \hspace{1cm} + \mathbbm{1}_{\{0\}}(\mathbf{L}_v[i,k]) \times 
		\frac{\partial \mathbf{L}_v[i,k]}{\partial \mathbf{v}[k]}
		\Big)~\text{(due to {\bf (a)})}
	\end{align*}
	\begin{align*}
		=& \frac{1}{m}
		\sum\limits_{\substack{i\in S\\\text{clause $i$ contains}\\\text{literal $p_k$}}}
		\Big(
		\mathbbm{1}_{\{1\}}(\mathbf{L}_v[i,k]) \times 
		\frac{\partial (1-\mathbf{L}_v[i,k])}{\partial \mathbf{v}[k]}
		\\
		& \hspace{1cm} + \mathbbm{1}_{\{0\}}(\mathbf{L}_v[i,k]) \times 
		\frac{\partial \mathbf{L}_v[i,k]}{\partial \mathbf{v}[k]}
		\Big) + \\
		& \frac{1}{m}
		\sum\limits_{\substack{i\in S\\\text{clause $i$ contains}\\\text{literal $\neg p_k$}}}
		\Big(
		\mathbbm{1}_{\{1\}}(\mathbf{L}_v[i,k]) \times 
		\frac{\partial (1-\mathbf{L}_v[i,k])}{\partial \mathbf{v}[k]}
		\\
		& \hspace{1cm} + \mathbbm{1}_{\{0\}}(\mathbf{L}_v[i,k]) \times 
		\frac{\partial \mathbf{L}_v[i,k]}{\partial \mathbf{v}[k]}
		\Big)~\text{(due to {\bf (b)})}
	\end{align*}
	\begin{align*}
		=& \frac{1}{m}
		\sum\limits_{\substack{i\in S\\\text{clause $i$ contains}\\\text{literal $p_k$}}}
		\Big(
		-\mathbbm{1}_{\{1\}}(\mathbf{L}_v[i,k])
		+ \mathbbm{1}_{\{0\}}(\mathbf{L}_v[i,k])
		\Big)\\
		& +\frac{1}{m}
		\sum\limits_{\substack{i\in S\\\text{clause $i$ contains}\\\text{literal $\neg p_k$}}}
		\Big(
		\mathbbm{1}_{\{1\}}(\mathbf{L}_v[i,k])
		- \mathbbm{1}_{\{0\}}(\mathbf{L}_v[i,k])
		\Big)\\
		& ~\text{(due to {\bf (c)} and {\bf (d)})}
	\end{align*}
}%
Recall that $\mathbf{L}_v[i,j]\in\{0,1\}$, and $\mathbf{L}_v[i,j]=1$ iff clause $i$ contains a literal ($p_j$ or $\neg p_j$) for atom $p_j$ and this literal evaluates to $\true$ under $v$.
It's easy to check that
\begin{itemize}
	\item when clause $i$ contains literal $p_k$, the value of $-\mathbbm{1}_{\{1\}}(\mathbf{L}_v[i,k])
	+ \mathbbm{1}_{\{0\}}(\mathbf{L}_v[i,k])$ is $-1$ if $v\models p_k$ and is $1$ if $v\not\models p_k$;
	\item when clause $i$ contains literal $\neg p_k$, the value of $\mathbbm{1}_{\{1\}}(\mathbf{L}_v[i,k])
	- \mathbbm{1}_{\{0\}}(\mathbf{L}_v[i,k])$ is $-1$ if $v\models p_k$ and is $1$ if $v\not\models p_k$.
\end{itemize}
Thus
\begin{align*}
	\frac{\partial L_{sat}}{\partial \mathbf{v}[k]} =& 
	\begin{cases}
		-\frac{c}{m} & \text{if $v\models p_k$,}\\
		\frac{c}{m} & \text{if $v\not \models p_k$.}
	\end{cases}
\end{align*}
where $c$ is the number of clauses in $S$ that contain a literal for atom $p_k$.
Finally, if $p_k\not \in F$,
\begin{align*}
	\frac{\partial L_{sat}}{\partial \xx[k]} 
	\stackrel{iSTE}{\approx}  \frac{\partial L_{sat}}{\partial \mathbf{v}[k]}
	=
	\begin{cases}
		-\frac{c}{m} & \text{if $v\models p_k$,} \\
		\frac{c}{m} & \text{if $v\not \models p_k$;} 
	\end{cases}
\end{align*}
where $c$ is the number of clauses in $C$ that are satisfied by $v$ and contain $p_k$ or $\neg p_k$.
\end{enumerate}
\end{proof}

\section{More Details about Experiments} \label{appendix:sec:cnf}

\subsection{mnistAdd2}
In {\bf mnistAdd2} problem \cite{manhaeve18deepproblog}, a data instance is a 5-tuple $\langle i_1, i_2, i_3, i_4, l \rangle$ such that $i_*$ are images of digits and $l$ is an integer in $\{0,\dots,198 \}$ denoting the sum of two 2-digit numbers $i_1i_2$ and $i_3i_4$.
The task is, given 15k data instances of $\langle i_1, i_2, i_3, i_4, l \rangle$, to train a CNN for digit classification given such weak supervision.
The CNF for {\bf mnistAdd2} consists of the 199 clauses of the form
\begin{align*}
\neg sum(l) \lor \bigvee\limits_{\substack{n_1,n_2,n_3,n_4\in \{0,\dots,9 \}:\\10(n_1+n_3)+n_2+n_4=l}} pred(n_1, n_2,n_3,n_4)
\end{align*}
for $l \in \{0,\dots,198\}$. Intuitively, this clause says that ``if the sum of $i_1i_2$ and $i_3i_4$ is $l$, then their individual labels $n_1,n_2,n_3,n_4$ must satisfy $10(n_1+n_3)+n_2+n_4=l$.''

This CNF contains $199$ clauses and $10^4+199=10199$ atoms for $pred/4$ and $sum/1$, respectively. 
According to the definition, we can construct the matrix $\mathbf{C} \in \{-1,0,1 \}^{199\times 10199}$ where each row represents a clause. 

To construct $\mathbf{f}$ and $\mathbf{v}$ for a data instance $\langle i_1,i_2,i_3,i_4,l \rangle$, the facts $\mathbf{f}$ is simply a vector in $\{0,1\}^{10199}$ with 10198 $0$s and a single $1$ for atom $sum(l)$; while the prediction $\mathbf{v}$ is a vector in $\{0,1\}^{10199}$ obtained as follows. We 
(i) feed images $i_1$,$i_2$,$i_3$,$i_4$ into the CNN and obtain the outputs $\xx_1, \xx_2, \xx_3, \xx_4 \in \mathbb{R}^{10}$ (consisting of probabilities);
(ii) construct $\mathbf{x} \in \mathbb{R}^{10000}$ such that its $(1000a+100b+10c+d)$-th element is $\xx_1[a] \times \xx_2[b] \times \xx_3[c] \times \xx_4[d]$ for $a,b,c,d \in \{0,\dots,9\}$; and
(iii) $\mathbf{v} = \mathbf{f} + \mathbbm{1}_{\{0 \}}(\mathbf{f}) \odot b_p(\xx)$. 


The loss function used for {\bf mnistAdd2} problem is
\begin{align*}
{\cal L} = \alpha L_{cnf}(\mathbf{C}, \mathbf{v}, \mathbf{f})  +\sum\limits_{\xx\in\{\xx_1,\dots,\xx_4 \}}\beta L_{bound}(\xx)
\end{align*}
where $\alpha=1$ and $\beta=0.01$.

\subsection{mnistAdd using b(x) and iSTE}
In {\bf mnistAdd} problem, a data instance is a 3-tuple $\langle i_1,i_2,l \rangle$ where $i_1, i_2$ are 2 images of digits and $l$ is an integer in $\{0,\dots,18\}$ indicating the sum of the 2 digit images. 
The propositional signature $\sigma$ in this problem consists of $139$ atoms:
$19$ atoms of the form $sum(i_1,i_2,s)$ for $s\in\{0,\dots,18\}$, 
$20$ atoms of the form $digit(i,n)$ for $i\in \{i_1,i_2\}$ and for $n\in\{0,\dots,9\}$, 
and $100$ atoms of the form $conj(i_1, n_1, i_2, n_2)$ for $n_1, n_2 \in \{ 0,\dots, 9\}$ (denoting the conjunction of $digit(i_1,n_1)$ and $digit(i_2,n_2)$). The CNF for this problem consists of $111$ clauses: 
$19$ clauses of the form 
\begin{align} \label{clause:mnistAdd:sum}
\neg sum(i_1,i_2,s) \lor \bigvee\limits_{\substack{n_1,n_2 \in \{0,\dots,9 \} \\ n_1+n_2=s}} conj(i_1, n_1, i_2, n_2)
\end{align}
for $s\in\{0,\dots,18\}$, 
$2$ clauses of the form 
\begin{align} \label{clause:mnistAdd:ec}
digit(i,0) \lor \dots \lor digit(i,9)
\end{align}
for $i\in \{i_1,i_2\}$, and 
$90$ clauses of the form
\begin{align} \label{clause:mnistAdd:uc}
\neg digit(i,n_1) \lor \neg digit(i,n_2)
\end{align}
for $i\in \{i_1,i_2\}$ and for $n_1, n_2 \in \{ 0,\dots, 9\}$ such that $n_1<n_2$. Intuitively, clause~\eqref{clause:mnistAdd:sum} says that ``if the sum of $i_1$ and $i_2$ is $s$, then we should be able to predict the labels $n_1, n_2$ of $i_1, i_2$ such that they sum up to $s$.'' Clauses \eqref{clause:mnistAdd:ec} and \eqref{clause:mnistAdd:uc} define the existence and uniqueness constraints on the label of $i$. 
Note that clauses \eqref{clause:mnistAdd:ec} and \eqref{clause:mnistAdd:uc} are not needed if we use $b_p(x)$+iSTE since these constraints will be enforced by the softmax function in the last layer of the neural network, which is widely and inherently used in most neuro-symbolic formalisms.

This CNF can be represented by the matrix $\mathbf{C}\in\{-1,0,1 \}^{111\times 139}$. 
To construct $\mathbf{f}$ and $\mathbf{v}$ for a data instance $\langle i_1,i_2,l \rangle$, the facts $\mathbf{f}$ is simply a vector in $\{0,1\}^{139}$ with 138 $0$s and a single $1$ for atom $sum(i_1,i_2,l)$; while the prediction $\mathbf{v}$ is a vector in $\{0,1\}^{139}$ obtained as follows. We 
(i) feed images $i_1$,$i_2$ into the CNN and obtain the outputs $\xx_1, \xx_2 \in \mathbb{R}^{10}$ (consisting of probabilities);
(ii) construct $\mathbf{x} \in \mathbb{R}^{139}$ such that its $(10a+b)$-th element is $\xx_1[a] \times \xx_2[b]$ for $a,b \in \{0,\dots,9\}$ and its remaining elements are $0$; and
(iii) $\mathbf{v} = \mathbf{f} + \mathbbm{1}_{\{0 \}}(\mathbf{f}) \odot b_p(\xx)$. 

Then, the total loss is defined as
\begin{align*}
{\cal L} = \alpha L_{cnf}(\mathbf{C}, \mathbf{v}, \mathbf{f})  +\sum\limits_{\xx\in\{\xx_1,\xx_2 \}}\beta L_{bound}(\xx)
\end{align*}
where $\alpha=1$ and $\beta=0.1$.

\subsection{add2x2}

In {\bf add2x2} problem, a data instance is a 8-tuple $\langle i_1,i_2,i_3,i_4,$ $row_1,row_2,col_1,col_2 \rangle$ where $i_*$ are 4 images of digits arranged in the following order in a grid
\begin{align*}
i_1 ~&~ i_2 \\
i_3 ~&~ i_4 ~,
\end{align*}
and each $row_*$ or $col_*$ is an integer in $\{0,\dots,18 \}$ denoting the sum of 2 digits on the specified row/column in the above grid.
The task is to train a CNN for digit classification given such weak supervision.

For $o,o' \in \{ (i_1, i_2), (i_3, i_4), (i_1, i_3), (i_2, i_4)\}$, and for $r \in \{ 0,\dots,18\}$ the CNF contains the following clause:
\begin{align*}
\neg sum(o,o',r) \lor \bigvee\limits_{i,j \in \{0,\dots,9 \}\atop i+j=r} conj(o, i, o', j). 
\end{align*}
This clause can be read as ``if the sum of $o$ and $o'$ is $r$, then $o$ and $o'$ must be some values $i$ and $j$ such that $i+j=r$.''
This CNF contains $4\times 19=76$ clauses and $76+ 4\times 10\times 10=476$ atoms (for $sum/3$ and $conj/4$, resp.). 
This CNF can be represented by the matrix $\mathbf{C} \in \{ -1,0,1\}^{76\times 476}$.

To construct $\mathbf{f}$ and $\mathbf{v}$ for a data instance $\langle i_1,i_2,i_3,i_4,$ $row_1,row_2,col_1,col_2 \rangle$, the facts $\mathbf{f}$ is simply a vector in $\{0,1\}^{476}$ with 472 $0$s and four $1$s for atoms $sum(i_1,i_2,row_1)$, $sum(i_3,i_4,row_2)$, $sum(i_1,i_3,col_1)$, and $sum(i_2,i_4,col_2)$; while the prediction $\mathbf{v}$ is a vector in $\{0,1\}^{476}$ obtained as follows. We 
(i) feed images $i_1$,$i_2$,$i_3$,$i_4$ into the CNN and obtain the outputs $\xx_1, \xx_2, \xx_3, \xx_4 \in \mathbb{R}^{10}$ (consisting of probabilities);
(ii) construct $\mathbf{x} \in \mathbb{R}^{476}$ as the concatenation of $\langle \mathbf{v}_1, \mathbf{v}_2, \mathbf{v}_3, \mathbf{v}_4, \{0\}^{76} \rangle$ where
\begin{align*}
\mathbf{v}_1 = \xx_1^T \cdot \xx_2, &\hspace{1cm} 
\mathbf{v}_2 = \xx_3^T \cdot \xx_4,\\
\mathbf{v}_3 = \xx_1^T \cdot \xx_3, &\hspace{1cm} 
\mathbf{v}_4 = \xx_2^T \cdot \xx_4;
\end{align*}
and
(iii) $\mathbf{v} = \mathbf{f} + \mathbbm{1}_{\{0 \}}(\mathbf{f}) \odot b_p(\xx)$. 

Then, the total loss is defined as
\begin{align*}
{\cal L} = \alpha L_{cnf}(\mathbf{C}, \mathbf{v}, \mathbf{f})  +\sum\limits_{\xx\in\{\xx_1,\dots,\xx_4 \}}\beta L_{bound}(\xx)
\end{align*}
where $\alpha=1$ and $\beta=0.1$.

\subsection{apply2x2}

In {\bf apply2x2} problem, a data instance is a 11-tuple $\langle d_1$, $d_2$, $d_3$, $o_{11}$, $o_{12}$, $o_{21}$, $o_{22}$, $row_1$, $row_2$, $col_1$, $col_2 \rangle$ where
$d_*$ are digits in $\{0,\dots,9\}$,
$o_*$ are 4 images of operators in $\{+,-,\times\}$ arranged in the following order in a grid
\begin{align*}
o_{11} ~&~ o_{12} \\
o_{21} ~&~ o_{22} ~,
\end{align*}
and each $row_*$ or $col_*$ is an integer denoting the value of the formula (e.g., $(4 \times 7) - 9$)
\begin{align} \label{formula:apply2x2}
(d_1~o_1~d_2)~o_2~d_3
\end{align}
where $(o_1,o_2)\in \{ (o_{11},o_{12}), (o_{21},o_{22}), (o_{11},o_{21}),$ $(o_{12},o_{22}) \}$ denotes the two operators on the specified row/column in the above grid.
The task is to train a CNN for digit classification given such weak supervision.

We construct a CNF to relate formula~\eqref{formula:apply2x2} and its value and will apply the CNF loss for $(o_1,o_2)\in \{$ $(o_{11},o_{12})$, $(o_{21},o_{22})$, $(o_{11},o_{21})$, $(o_{12},o_{22}) \}$.

For $d_1, d_2, d_3 \in \{ 0,\dots,10\}$, and for all possible $r$ such that $(d_1~ Op_1~ d_2)~ Op_2~ d_3=r$ for some $Op_1, Op_2\in \{+,-, \times\}$, the CNF contains the following clause:
\begin{align*}
&\neg apply(d_1, o_1, d_2, o_2, d_3, r) \lor \\ &\bigvee\limits_{Op_1, Op_2\in \{+,-, \times\}\atop (d_1~ Op_1~ d_2)~ Op_2~ d_3=r} (operators(o_1, Op_1, o_2, Op_2)).
\end{align*}
This clause can be read as ``if the result is $r$ after applying $o_1$ and $o_2$ to the three digits, then $o_1$ and $o_2$ must be some values $Op_1$ and $Op_2$ such that $(d_1~ Op_1~ d_2)~ Op_2~ d_3=r$.''
This CNF contains $10597$ clauses and $10606$ atoms and can be represented by the matrix $\mathbf{C} \in \{ -1,0,1\}^{10597\times 10606}$. 

Given a data instance $\langle d_1$, $d_2$, $d_3$, $o_{11}$, $o_{12}$, $o_{21}$, $o_{22}$, $row_1$, $row_2$, $col_1$, $col_2 \rangle$, we construct $\mathbf{v}_i, \mathbf{f}_i\in \{0,1 \}^{10606}$ for $i\in \{1,\dots,4\}$, one for each $\langle o_1, o_2, r \rangle \in \{ \langle o_{11}, o_{12}, row_1 \rangle,$ $ \langle o_{21}, o_{22}, row_2 \rangle, \langle o_{11}, o_{21}, col_1 \rangle, \langle o_{12}, o_{22}, col_2 \rangle \}$. The detailed steps to construct $\mathbf{f}$ and $\mathbf{v}$ for $\langle o_1, o_2, r \rangle$ is as follows.

First, the facts $\mathbf{f}$ is simply a vector in $\{0,1\}^{10606}$ with 10605 $0$s and a single $1$ for atom $apply(d_1,o_1,d_2,o_2,d_3,r)$. Second, the prediction $\mathbf{v}$ is a vector in $\{0,1\}^{10606}$ obtained as follows. We 
(i) feed images $o_1$,$o_2$ into the CNN and obtain the outputs $\xx_1, \xx_2 \in \mathbb{R}^{3}$ (consisting of probabilities);
(ii) construct $\mathbf{x} \in \mathbb{R}^{10606}$ such that its $(3a+b)$-th element is $\xx_1[a] \times \xx_2[b]$ for $a,b \in \{0,\dots,2\}$ and its remaining elements are $0$; and
(iii) $\mathbf{v} = \mathbf{f} + \mathbbm{1}_{\{0 \}}(\mathbf{f}) \odot b_p(\xx)$. 

Then, the total loss is defined as
\begin{align*}
{\cal L} = \sum\limits_{i\in\{1,\dots,4 \}} \alpha L_{cnf}(\mathbf{C}, \mathbf{v}_i, \mathbf{f}_i)  +\sum\limits_{\xx\in\{\xx_1,\dots,\xx_4 \}}\beta L_{bound}(\xx)
\end{align*}
where $\alpha=1$ and $\beta=0.1$.


\subsection{member(n)}

We take {\bf member(3)} problem as an example.
In {\bf member(3)} problem, a data instance is a 5-tuple $\langle i_1,i_2,i_3,d,l \rangle$ where $i_1, i_2, i_3$ are 3 images of digits, $d$ is a digit in $\{0,\dots,9 \}$, and $l$ is an integer in $\{0,1\}$ indicating whether $d$ appears in the set of digit images. 
The task is to train a CNN for digit classification given such weak supervision.
The CNF for this problem consists of the 2 kinds of clauses in table~\ref{table:member3_cnf_example}.

\begin{table}[ht]
\centering
\caption{Clauses in the CNF for {\bf member(3)} Problem}
\label{table:member3_cnf_example}
\begin{tabular}{ |p{0.50\linewidth}|p{0.4\linewidth}| } 
\hline
Clause & Reading \\ 
\hline
$\neg in(d,1) \lor digit(i_1, d) \lor digit(i_2, d) \lor digit(i_3, d)$ \newline (for $d \in \{0,\dots,9\}$) & if $d$ appears in the 3 images, then $i_1$ or $i_2$ or $i_3$ must be digit $d$  \\ 
\hline
$\neg in(d,0) \lor \neg digit(i, d)$ \newline (for $d \in \{0,\dots,9\}$ and $i\in \{i_1,i_2,i_3\}$) & if $d$ doesn't appear in the 3 images, then each image $i$ cannot be digit $d$  \\ 
\hline
\end{tabular}
\end{table}

This CNF contains $10 + 10\times 3 =40$ clauses and $3\times 10 + 2\times 10 = 50$ atoms for $digit/2$ and $in/2$ respectively. According to the definition, we can construct the matrix $\mathbf{C} \in \{-1,0,1 \}^{40\times 50}$ where each row represents a clause. For instance, the row for the clause $\neg in(5,1) \lor digit(i_1, 5) \lor digit(i_2, 5) \lor digit(i_3, 5)$ is a row vector in $\{-1,0,1\}^{1\times 50}$ containing 46 $0$s, a single $-1$ for atom $in(5,1)$, and three $1$s for atoms $digit(i_1, 5), digit(i_2, 5), digit(i_3, 5)$.

To construct $\mathbf{f}$ and $\mathbf{v}$ for a data instance $\langle i_1,i_2,i_3,d,l \rangle$, the facts $\mathbf{f}$ is simply a vector in $\{0,1\}^{50}$ with 49 $0$s and a single $1$ for atom $in(d,l)$; while the prediction $\mathbf{v}$ is a vector in $\{0,1\}^{50}$ obtained as follows. We 
(i) feed images $i_1$,$i_2$,$i_3$ into the CNN and obtain the NN outputs $\xx_1, \xx_2, \xx_3 \in \mathbb{R}^{10}$ consisting of probabilities,
(ii) construct $\xx \in \mathbb{R}^{50}$ by concatenating $\xx_1, \xx_2, \xx_3$ and the vector $\{0\}^{20}$, and
(iii) $\mathbf{v} = \mathbf{f} + \mathbbm{1}_{\{0 \}}(\mathbf{f}) \odot b_p(\xx)$.

The total loss function is
\begin{align*}
{\cal L} = \alpha L_{cnf}(\mathbf{C}, \mathbf{v}, \mathbf{f})  +\sum\limits_{\xx\in\{\xx_1,\dots,\xx_3 \}}\beta L_{bound}(\xx)
\end{align*}
where $\alpha=1$ and $\beta=0.1$.

\subsection{Sudoku} \label{appendix:subsec:cnf-sudoku}

We use a typical CNF for $9\times 9$ Sudoku problem. The CNF is defined on a propositional signature $\sigma=\{a(R,C,N) \mid R,C,N \in \{1,\dots,9\} \}$ where $a(R,C,N)$ represents ``digit $N$ is assigned at row $R$ column $C$''. The CNF consists of the following $1+\binom{9}{2}=37$ clauses for each of the $4\times 9\times 9=324$ different sets $A$ of atoms
\begin{align*}
&\bigvee\limits_{p\in A} p \\
&\neg p_i \lor \neg p_j  ~~~(\text{for $p_i, p_j \in A$ and $i<j$})
\end{align*}
where the $4\times 9\times 9$ definitions of $A$ can be split into the following $4$ categories, each consisting of $9\times 9$ definitions.
\begin{enumerate}
\item (UEC on row indices) \newline
For $C,N\in \{1,\dots,9\}$, $A$ is the set of atoms $\{a(1,C,N), \dots, a(9,C,N)\}$.
\item (UEC on column indices) \newline 
For $R,N\in \{1,\dots,9\}$, $A$ is the set of atoms $\{a(R,1,N), \dots, a(R,9,N)\}$.
\item (UEC on 9 values in each cell) \newline 
For $R,C\in \{1,\dots,9\}$, $A$ is the set of atoms $\{a(R,C,1), \dots, a(R,C,9)\}$.
\item (Optional: UEC on 9 cells in the same $3\times 3$ box) \newline 
For $B,N\in \{1,\dots,9\}$, $A$ is the set of atoms $\{a(R_1,C_1,N), \dots, a(R_9,C_9,N)\}$ such that the 9 cells $(R_i,C_i)$ for $i\in \{1\dots,9 \}$ are the 9 cells in the $B$-th box in the $9\times 9$ grid for value $N$. Note that the clauses in bullet 4 are optional under the setting $b_p(x)$+iSTE since they are already enforced by the softmax function used in the last layer to turn NN output to probabilities.

This CNF can be represented by a matrix $\mathbf{C}\in\{-1,0,1 \}^{8991\times 729}$. 
The dataset in the CNN experiments consists of 70k data instances, 
20\% supervised for testing, and 80\% unsupervised for training.
Each unsupervised data instance is a single vector $\mathbf{q}\in \{0,1,\dots,9 \}^{81}$ representing a $9\times 9$ Sudoku board ($0$ denotes an empty cell).
The non-zero values in $\mathbf{q}$ are treated as atomic facts $F$ and we construct the matrix $\mathbf{F} \in\{0,1\}^{81\times 9}$ such that, for $i\in \{ 1,\dots,81\}$, the $i$-th row $\mathbf{F}[i,:]$ is the vector $\{0\}^9$ if $\mathbf{q}[i]=0$ and is the one-hot vector for $\mathbf{q}[i]$ if $\mathbf{q}[i]\neq 0$. Then, the vector $\mathbf{f} \in\{0,1\}^{729}$ is simply the flattened version of $\mathbf{F}$. We feed $\mathbf{q}$ into the CNN and obtain the output $\xx\in \mathbb{R}^{729}$ consisting of probabilities.  
The prediction $\mathbf{v}\in\{0,1\}^{729}$ is obtained as $\mathbf{f} + \mathbbm{1}_{\{0 \}}(\mathbf{f}) \odot b_p(\xx)$.

Then, the total loss function ${\cal L}$ used to train the CNN for Sudoku is
\begin{align*}
{\cal L} = \alpha L_{cnf}(\mathbf{C}, \mathbf{v}, \mathbf{f}) + \beta L_{bound}(\xx)
\end{align*}
where $\alpha=1$ and $\beta=0.1$.
\end{enumerate}

\section{Ablation Study with Sudoku-GNN} \label{appendix:sudoku:gnn}
To better analyze the effect of constraint losses on general GNN, in this section, we apply constraint losses to a publicly available GNN for Sudoku problem.%
\footnote{
The GNN is from https://www.kaggle.com/matteoturla/can-graph-neural-network-solve-sudoku, along with the dataset. 
}
The graph for Sudoku problem consists of 81 nodes, one for each cell in the Sudoku board, and 1620 edges, one for each pair of nodes in the same row, column, or $3\times 3$ non-overlapping box. 
The GNN consists of an embedding layer, 8 iterations of message passing layers, and an output layer. 

For each data instance $\langle \mathbf{q}, \mathbf{l}\rangle$, the GNN takes $\mathbf{q}\in \{0,1,\dots,9 \}^{81}$ as input and outputs a matrix of probabilities $\mathbf{X} \in \mathbf{R}^{81\times 9}$ after 8 message passing steps. 

The baseline loss $L_{baseline}$ is the cross-entropy loss defined on prediction $\mathbf{X}$ and label $\mathbf{l}$.
\begin{align*}
L_{baseline} = L_{cross\_entropy}(\mathbf{X}, \mathbf{l})
\end{align*}
The constraint loss $L_{cl}$ is the same as the total loss in Appendix~\ref{appendix:subsec:cnf-sudoku} where $\xx$ is the flattening of $\mathbf{X}$.
\begin{align}
L_{cl} = L_{cnf}(\mathbf{C}, \mathbf{v}, \mathbf{f}) + 0.1\times L_{bound}(\xx).  \label{eq:appendix:L_cl:sudoku}
\end{align}

In addition, we designed the following domain-specific loss functions for Sudoku problem as semantic regularizers for comparison. Intuitively, $L_{hint}$ says that ``the given digits must be predicted'' and $L_{sum}$ says that ``the sum of the 9 probabilities in $\mathbf{X}$ in the same row/column/box must be 1''.
\begin{align*}
L_{hint} &= avg\Big(\mathbf{f} \odot (1-b_p(\mathbf{x}))\Big)\\
L_{sum} &= \sum\limits_{s\in \{1,\dots,32 \}\atop i\in \{row,col,box\}} avg((sum(\mathbf{X}_s^i) - 1)^2). 
\end{align*}
Here, $avg(X)$ and $sum(X)$ compute the average and sum of all elements in $X$ along its last dimension;
$\mathbf{X}_s^{row}, \mathbf{X}_s^{col}, \mathbf{X}_s^{box} \in \mathbb{R}^{81\times 9}$ are reshaped copies of $\mathbf{X}_s$ such that each row in, for example, $\mathbf{X}_s^{row}$ contains 9 probabilities for atoms $a(1,C,N), \dots, a(9,C,N)$ for some $C$ and $N$. 

\begin{figure}[ht!]
\begin{center}
\includegraphics[width=0.7\columnwidth]{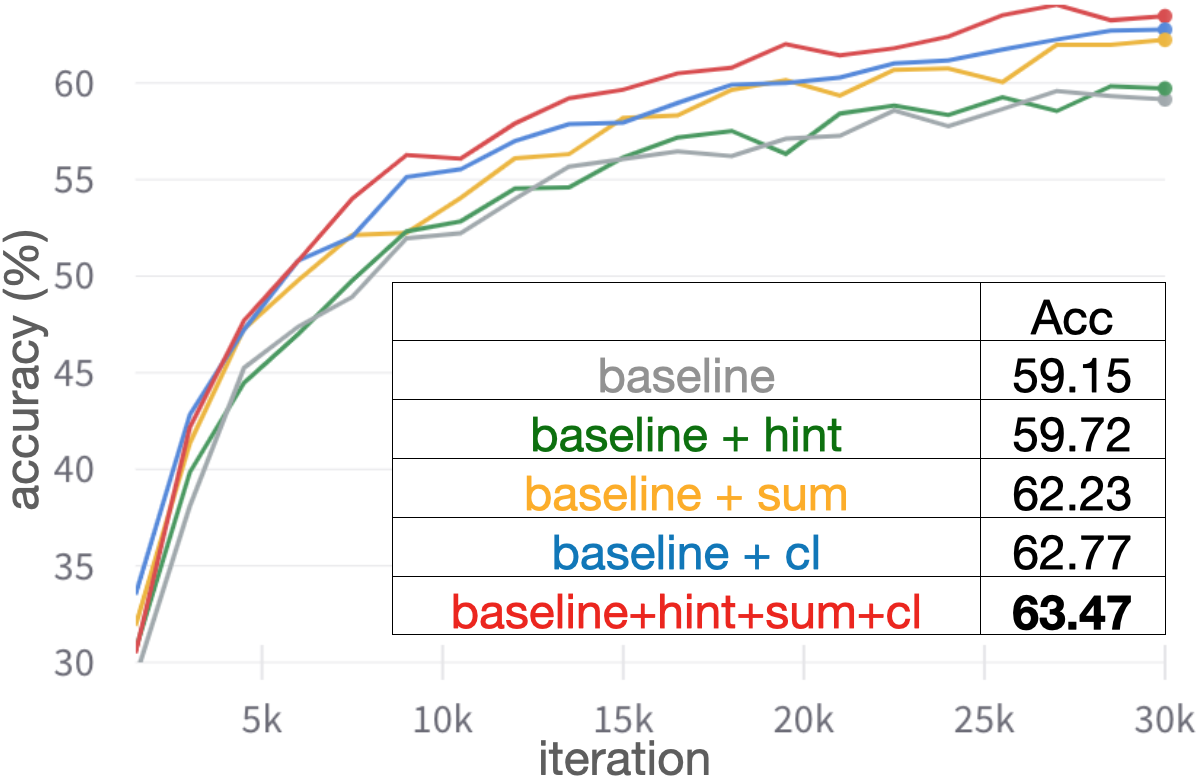}
\vspace{-3mm}
\caption{Acc with 30k dataset under different losses}
\label{fig:sudoku:gnn:30k:sup:20epochs:stecnf}
\vspace{-3mm}
\end{center}
\end{figure}

\begin{figure}[ht!]
\begin{center}
\includegraphics[width=0.7\columnwidth]{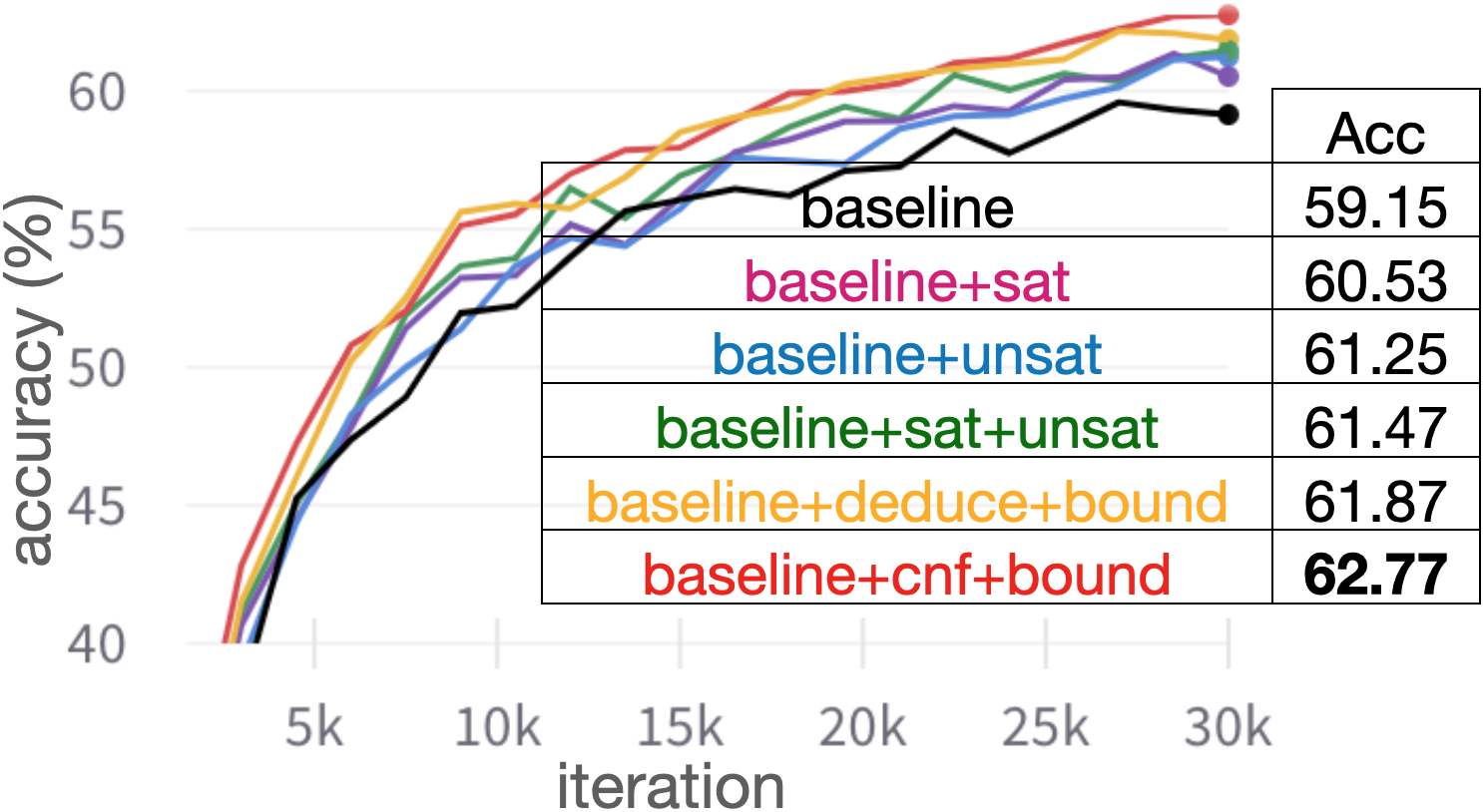}
\vspace{-3mm}
\caption{Acc with 30k dataset under different losses in $L_{cl}$}
\label{fig:sudoku:gnn:30k:sup:20epochs:L_ste}
\vspace{-3mm}
\end{center}
\end{figure}

Figure~\ref{fig:sudoku:gnn:30k:sup:20epochs:stecnf} shows the test accuracy of the GNN after 20 epochs of training on 30k data instances (with full supervision) using different loss functions (denoted by subscripts of losses). 
It shows monotonic improvement from each loss and the best result is achieved when we add all losses. 

Figure~\ref{fig:sudoku:gnn:30k:sup:20epochs:L_ste} further shows the monotonic improvement from each component in
\begin{align*}
L_{cl}=L_{deduce} + L_{sat} + L_{unsat} + 0.1\times L_{bound}
\end{align*}
where we split $L_{cnf}(\mathbf{C}, \mathbf{v}, \mathbf{f})$ in equation~\eqref{eq:appendix:L_cl:sudoku} into its 3 components.
We can see that the most improvement comes from {$L_{deduce} + 0.1\times L_{bound}$}, which aligns with Proposition~\ref{prop:gradient:b_p} since $L_{deduce}$ has dominant gradients that enforces a deduction step. Noticeably, $L_{bound}$ is necessary for $L_{deduce}$ to bound the size of raw NN output.

\begin{figure}[ht!]
\begin{center}
\centerline{\includegraphics[width=0.7\columnwidth]{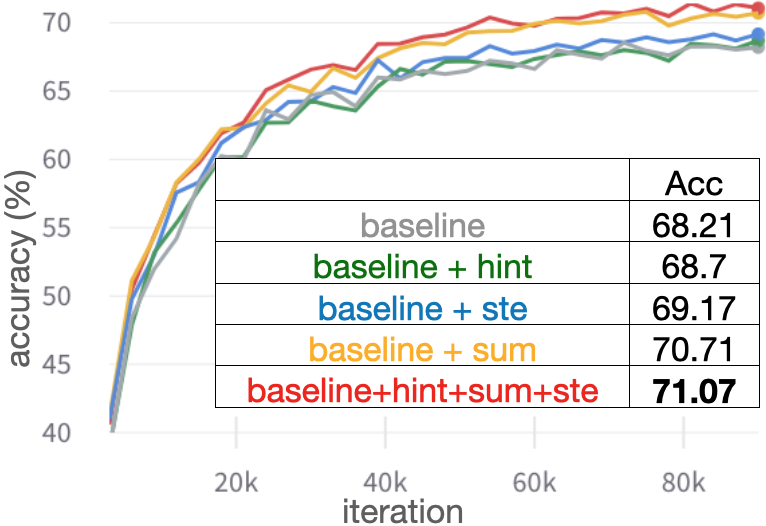}}
\vspace{-3mm}
\caption{Acc with 60k dataset under different losses}
\label{fig:sudoku:gnn:60k:sup:60epochs:stecnf}
\vspace{-3mm}
\end{center}
\end{figure}

Figure~\ref{fig:sudoku:gnn:60k:sup:60epochs:stecnf} shows the test accuracy of the GNN after 60 epochs of training on 60k data instances (with full supervision). 
We can see that the monotonic improvement from each loss is kept in the experiments with 60k data instances and the best result is still achieved when we add all losses. 
However, the most improvement is from $L_{sum}$ instead of $L_{cl}$. This is because most semantic information in $L_{cl}$ are from $L_{deduce}$ (i.e., one step deduction from the given digits), which can be eventually learned by the GNN with more data instances.



\section{More Examples} \label{sec:more-examples}

\subsection{Learning to Solve the Shortest Path Problem} \label{ssec:shortest-path}
The experiment is about, given a graph and two points, finding the shortest path between them.
We use the dataset from \cite{xu18asemantic}, which was used to demonstrate the effectiveness of semantic constraints for enhanced neural network learning. The dataset is divided into 80/20 train/test examples. Each example is a 4 by 4 grid $G = (V, E)$, where $\lvert V \rvert = 16, \lvert E \rvert = 24$, two-terminal (i.e., source and the destination) nodes are randomly picked up from 16 nodes, and 8 edges are randomly removed from 24 edges to increase the difficulty. The dataset consists of 1610 data instances, each is a pair $\langle \mathbf{i}, \mathbf{l} \rangle$ where $\mathbf{i} \in \{0,1 \}^{40}$ and $\mathbf{l} \in \{0,1 \}^{24}$. The ones in the first 24 values in $\mathbf{i}$ denote the (non-removed) edges in the grid, the ones in the last 16 values in $\mathbf{i}$ denote the terminal nodes, and ones in $\mathbf{l}$ denote the edges in the shortest path.

We define a CNF with 40 atoms and 120 clauses to represent ``each terminal node is connected to exactly one edge in the shortest path''. 
To start with, let's identify each node in the $4\times 4$ grid by a pair $(i,j)$ for $i,j \in \{1,\dots,4\}$ and identity the edge between nodes $(i_1,j_1)$ and $(i_2,j_2)$ as $((i_1,j_1), (i_2,j_2))$. Then, we introduce the following 2 atoms.
\begin{itemize}
\item $terminal(i,j)$ represents that node $(i,j)$ is one of the two terminal nodes.
\item $sp((i_1,j_1), (i_2,j_2))$ represents edge $((i_1,j_1),(i_2,j_2))$ is in the shortest path.
\end{itemize}
Then, the CNF for the shortest path problem consists of 120 clauses: 16 clauses of the form
\begin{align*}
\neg terminal(i_1,j_1) \lor \bigvee\limits_{\substack{i_2,j_2:\\((i_1,j_1), (i_2,j_2))\\\text{is an edge}}}sp((i_1,j_1), (i_2,j_2))
\end{align*}
for $i_1,j_1\in \{1,\dots,4 \}$, and 104 clauses of the form
\begin{align*}
\neg terminal(i_1,j_1) &\lor \neg sp((i_1,j_1), (i_2,j_2))\\
&\lor \neg sp((i_1,j_1), (i_3,j_3))
\end{align*}
for $i_*,j_* \in \{1,\dots,4 \}$ such that $((i_1,j_1), (i_2,j_2))$ and $((i_1,j_1), (i_3,j_3))$ are different edges. 

This CNF can be represented by a matrix $\mathbf{C}\in\{-1,0,1 \}^{120\times 40}$. 

To construct $\mathbf{f}$ and $\mathbf{v}$ for a data instance $\langle \mathbf{i}, \mathbf{l} \rangle$, the facts $\mathbf{f} \in \{0,1\}^{40}$ is simply the concatenation of $\mathbf{i}[24:]$ and $\{0\}^{24}$; while the prediction $\mathbf{v}$ is a vector in $\{0,1\}^{40}$ obtained as follows. We 
(i) feed $\mathbf{i}$ into the same MLP from \cite{xu18asemantic} and obtain the NN output $\mathbf{x}\in [0,1]^{24}$ consisting of probabilities,
(ii) extend $\mathbf{x}$ with $16$ $0s$ (in the beginning) so as to have a 1-1 correspondence between $40$ elements in $\xx$ and $40$ atoms in the CNF, and
(iii) $\mathbf{v} = \mathbf{f} + \mathbbm{1}_{\{0 \}}(\mathbf{f}) \odot b_p(\xx)$.

Finally, the total loss function ${\cal L}_{base}$ used in the baseline is
\begin{align*}
{\cal L}_{base} &= L_{cross}(\xx, \mathbf{l})
\end{align*}
where $L_{cross}$ is the cross-entropy loss.

The loss function ${\cal L}$ used for shortest path problem is
\begin{align*}
{\cal L} &= L_{cross}(\xx, \mathbf{l}) + \alpha L_{cnf}(\mathbf{C}, \mathbf{v}, \mathbf{f}) + \beta L_{bound}(\xx)
\end{align*}
where $\alpha=0.2$ and $\beta=1$. 
We set $\alpha=0.2$ in our experiments to balance the gradients from the CNF loss and cross entropy loss. Indeed, a similar accuracy can be achieved if we compute $\alpha$ dynamically as $\frac{g_{cross}}{g_{cnf}}$ where $g_{cnf}$ and $g_{cross}$ are the maximum absolute values in the gradients $\frac{\partial L_{cnf}(\mathbf{C}, \mathbf{v}, \mathbf{f})}{\partial \xx}$ and $\frac{\partial L_{cross}(\xx, \mathbf{l})}{\partial \xx}$ respectively.
Intuitively, the weight $\alpha$ makes sure that the semantic regularization from CL-STE won't overwrite the hints from labels.

\begin{figure}[ht]
\begin{center}
\centerline{\includegraphics[width=0.8\columnwidth,height=3.5cm]{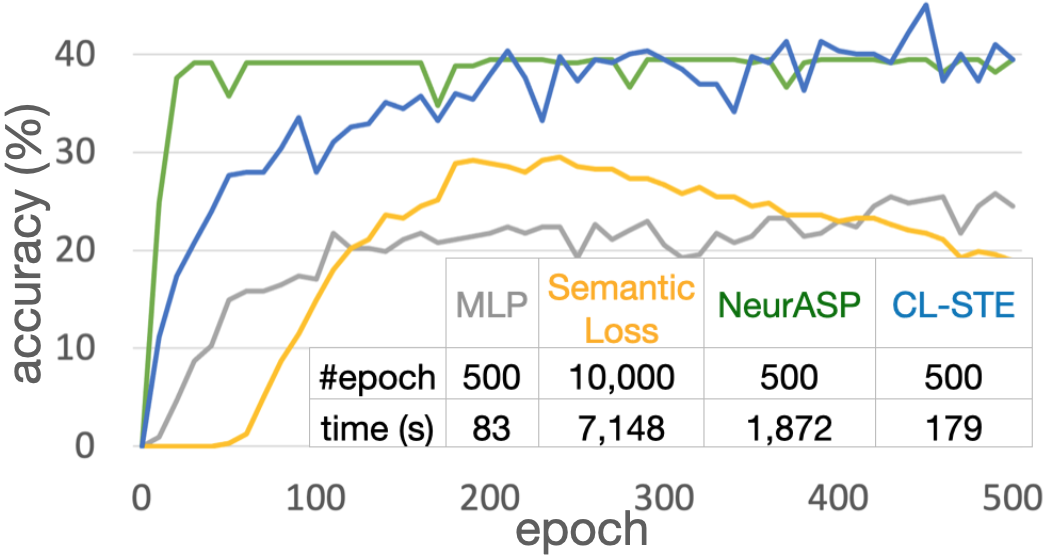}}
\caption{MLP+CL-STE on Shortest Path Problem}
\label{fig:tb:sp}
\vspace{-0.5cm}
\end{center}
\end{figure}
Figure~\ref{fig:tb:sp} compares the test accuracy of the same Multi-Layer Perceptron (MLP) trained by different learning methods during 500 epochs of training (except that the accuracy for Semantic Loss method is reported for 10k epochs).
As we can see, it only took 83s for baseline and 179s for CL-STE to complete all 500 epochs (including the time to compute training and testing accuracy) since they are all trained on GPU with a batch size of 32. 
Besides, CL-STE achieves comparable accuracy to NeurASP with only about $\frac{1}{10}$ of time. 
The training time of Semantic Loss in Figure~\ref{fig:tb:sp} was recorded when it was trained on CPU. We re-did the Semantic Loss experiment on GPU with early stopping and found that it still takes 1032s to achieve the highest accuracy 30.75\% after 2900 epochs of training.

\subsection{Semi-Supervised Learning for MNIST and FASHION Dataset} \label{sec:semi}
Xu {\sl et al.} (\citeyear{xu18asemantic}) show that minimizing semantic loss could enhance semi-supervised multi-class classification results by enforcing the constraint that a model must assign a unique label even for unlabeled data.
Their method achieves state-of-the-art results on the permutation invariant MNIST classification problem, a commonly used testbed for semi-supervised learning algorithms, and a slightly more challenging problem, FASHION-MNIST. 

For both tasks, we apply $b(x)$+iSTE to the same MLP (without softmax in the last layer) as in \cite{xu18asemantic}, i.e., an MLP of shape (784, 1000, 500, 250, 250, 250, 10), where the output $\xx \in \mathbb{R}^{10}$ denotes the digit/cloth prediction.

The CNF for this problem consists of $46$ clauses: 
$1$ clause
\begin{align*}
pred(i,0) \lor \dots \lor pred(i,9)
\end{align*}
and $45$ clauses of the form
\begin{align*}
\neg pred(i,n_1) \lor \neg pred(i,n_2)
\end{align*}
for $n_1, n_2 \in \{ 0,\dots, 9\}$ such that $n_1<n_2$. Intuitively, these 2 clauses define the existence and uniqueness constraints on the label of image $i$. 
This CNF can be represented by the matrix $\mathbf{C} \in \{ -1,0,1\}^{46\times 10}$. 

The vectors $\mathbf{f}$ and $\mathbf{v}$ are constructed in the same way for both supervised data instance $\langle i,l \rangle$ and unsupervised data instance $\langle i \rangle$.  The facts $\mathbf{f}$ is simply $\{0\}^{10}$; while the prediction $\mathbf{v}$ is a vector in $\{0,1\}^{10}$ obtained as follows. We 
(i) feed image $i$ into the CNN and obtain the outputs $\xx \in \mathbb{R}^{10}$ (consisting of real values not probabilities); and
(ii) $\mathbf{v} = \mathbf{f} + \mathbbm{1}_{\{0 \}}(\mathbf{f}) \odot b(\xx)$. 
Then, the total loss for unsupervised data instances is defined as
\begin{align*}
{\cal L} = L_{cnf}(\mathbf{C}, \mathbf{v}, \mathbf{f})  +L_{bound}(\xx),
\end{align*}
which enforces that each image should map to exactly one digit or one cloth type. The total loss for supervised data instance simply contains ${\cal L}$ as well as the typical cross-entropy loss.

We train the network using 100, 500, and 1,000 partially labeled data and full (60,000) labeled data, respectively.
We run experiments for 50k batch updates with a batch size of 32.
Each experiment is repeated 10 times, and we report the mean and the standard deviation of classification accuracy (\%). 

\begin{table}[ht]
\centering
{\scriptsize
\caption{Accuracy on MNIST \& FASHION dataset}
\label{tbl:mnist-semi}
\begin{tabular}{l|c|c|c|c} \toprule
	\multicolumn{1}{c}{\multirow{2}*{Method}} & \multicolumn{4}{|c}{Number of labeled examples used} \\ \cline{2-5}
	& 100 & 500 & 1000 & All (60,000)  \\ \hline
	MNIST (Xu et al.) & 85.3$\pm$1.1 & 94.2$\pm$0.5 & 95.8$\pm$0.2 & 98.8$\pm$0.1 \\ 
	\hline
	MNIST ($b(x)$+iSTE) & 84.4$\pm$1.5 & 94.1$\pm$0.3 & 95.9$\pm$0.2 & 98.8$\pm$0.1 \\ \bottomrule
	FASHION (Xu et al.) & 70.0$\pm$2.0 & 78.3$\pm$0.6 & 80.6$\pm$0.3 & 87.3$\pm$0.2 \\ 
	\hline
	FASHION ($b(x)$+iSTE) & 71.0$\pm$1.2 & 78.6$\pm$0.7 & 80.7$\pm$0.5 & 87.2$\pm$0.1 \\ 
	\bottomrule   
\end{tabular}
}
\end{table}

Table~\ref{tbl:mnist-semi} shows that the MLP with the CNF loss achieves similar accuracy with the implementation of semantic loss from~\cite{xu18asemantic}. 
Time-wise, each experiment using the method from \cite{xu18asemantic} took up about 12 minutes, and each experiment using the CL-STE method took about 10 minutes. There is not much difference in training time since the logical constraints for this task in the implementation of semantic loss \cite{xu18asemantic} are simple enough to be implemented in python scripts without constructing an arithmetic circuit and inference on it.